\documentclass[sigconf]{acmart}
\renewcommand\footnotetextcopyrightpermission[1]{}

\usepackage{multirow}
\usepackage{threeparttable}
\usepackage{color}
\usepackage{soul}
\usepackage{wrapfig}
\usepackage[most]{tcolorbox}  
\usepackage{enumitem}
\usepackage{tabularx}
\usepackage{cuted}
\usepackage{caption}
\usepackage{booktabs}

\newcommand{\name}{MobCache}
\AtBeginDocument{%
  }

\setcopyright{acmlicensed}
\copyrightyear{2018}
\acmYear{2018}
\acmDOI{XXXXXXX.XXXXXXX}
\acmConference[Conference acronym 'XX]{Make sure to enter the correct
  conference title from your rights confirmation email}{June 03--05,
  2018}{Woodstock, NY}
\acmISBN{978-1-4503-XXXX-X/2018/06}

\begin{document}

\title{Mobility-Aware Cache Framework for Scalable LLM-Based Human Mobility Simulation}

\author{Hua Yan}
\affiliation{%
  \institution{Lehigh University}
  \city{Bethlehem}
  \country{USA}}
\email{huy222@lehigh.edu
}
\author{Heng Tan}
\affiliation{%
  \institution{Lehigh University}
  \city{Bethlehem}
  \country{USA}}
\email{het221@lehigh.edu
}

\author{Yingxue Zhang}
\affiliation{%
  \institution{State University of New York at Binghamton}
  \city{Binghamton}
  \country{USA}}
\email{yzhang42@binghamton.edu
}

\author{Yu Yang}
\affiliation{%
  \institution{Lehigh University}
  \city{Bethlehem}
  \country{USA}}
\email{yuyang@lehigh.edu
}


\begin{abstract}
Simulating large-scale human mobility is fundamental to understanding population movement patterns and supporting real-world geospatial applications such as urban planning, epidemic response, and transportation analysis.
Recent works treat large language models (LLMs) as human agents to simulate realistic mobility behaviors using structured reasoning, but their high computational cost limits scalability.
To address this, we design a mobility-aware cache framework named \name{} that leverages reconstructible caches to enable efficient large-scale human mobility simulations.
It consists of: (1) a reasoning component that encodes each reasoning step as a latent-space embedding and uses a latent-space evaluator to enable the reuse and recombination of reasoning steps; and (2) a decoding component that employs a lightweight decoder trained with mobility law-constrained distillation to translate latent-space reasoning chains into natural language, thereby improving simulation efficiency while maintaining fidelity.
Experiments show that \name{} significantly improves efficiency across multiple dimensions while maintaining performance comparable to state-of-the-art LLM-based methods. 
\end{abstract}




\maketitle

\section{Introduction}

Human mobility modeling plays a fundamental role in supporting a wide range of downstream geospatial and spatiotemporal applications, including urban planning~\cite{wu2024metaurban,wu2025towards}, epidemiology~\cite{geng2022hmes,fan2020human}, and transportation analysis~\cite{tan2023joint,nooshi2025multi}.
Effective human mobility analysis typically relies on large-scale, fine-grained mobility data that capture population-level movement patterns over space and time. In practice, such data are mainly obtained through two channels. Travel surveys record individual trips but suffer from recall bias, sparse temporal sampling, and high collection costs~\cite{toch2019analyzing,bricka2024summary}. Sensor-based tracking, such as mobile-phone traces or Bluetooth beacons, provides denser temporal coverage but depends on device penetration and raises significant privacy concerns~\cite{chicago_data,lajoie2024peoplex}. As a result, both channels face challenges in scaling to millions of agents while preserving privacy, motivating the need for privacy-preserving alternatives such as mobility simulation.

A growing body of recent work leverages large language models (LLMs) to simulate human mobility trajectories without relying on real mobility traces~\cite{jiawei2024large,du2025cams,piao2025agentsociety,liu2024human,ju2025trajllm,shao2024chain,bhandari2024urban,li2024geo}, which show promising results.
These methods typically model an LLM as a virtual human agent and prompt it to perform step-by-step reasoning over mobility intentions and activities. While such methods can produce realistic mobility behaviors, they often incur substantial computational costs. For instance, simulating one million agents for a single day can exceed \$1,000 in API fees under current token pricing schemes~\cite{shah2025llmcost,chen2023frugalgpt}. 
There are two main methods to reduce costs.
First, a group-based methods divides agents into coarse profile groups (e.g., by job and income) and calls the LLM once per group to generate trajectories~\cite{chopra2024limits}. However, the method forces every agent in the same group to share identical behavior, which reduces individual diversity.
Second, by processing multiple I/O operations simultaneously and reusing TCP connections, the system can handle many requests at the same time~\cite{yan2024opencity,piao2025agentsociety}. But each agent still needs one LLM call for every trajectory, so the total monetary cost of these methods remains high.

A natural way to reduce repeated LLM calls is \emph{response caching}, where input--output query pairs are stored locally and reused whenever a similar query arises. 
This caching paradigm has proven effective in many domains such as e-commence~\cite{zhao2025explainable}, intelligent QA systems~\cite{couturier2025semantic} and Machine Translation~\cite{gim2024prompt}.
Note that this paradigm is orthogonal to other caching mechanisms such as KV caching~\cite{jin2024ragcache,zheng2023efficiently}, and we provide a comparison in the related work section.
However, response caching suppresses diversity. Common queries, such as \emph{``What is the next activity for a software engineer who has just finished work?''}, often hit the same cached answer, for example a nine‑to‑five schedule, even though real workers also take night shifts, overtime, or flexible hours. The resulting lack of behavioral variation lowers simulation fidelity (see Section~\ref{sec:cache_matter} and Section~\ref{sec:eva_group}).

\begin{figure}[t]
\centering
\includegraphics[width=0.85\linewidth, keepaspectratio=true]{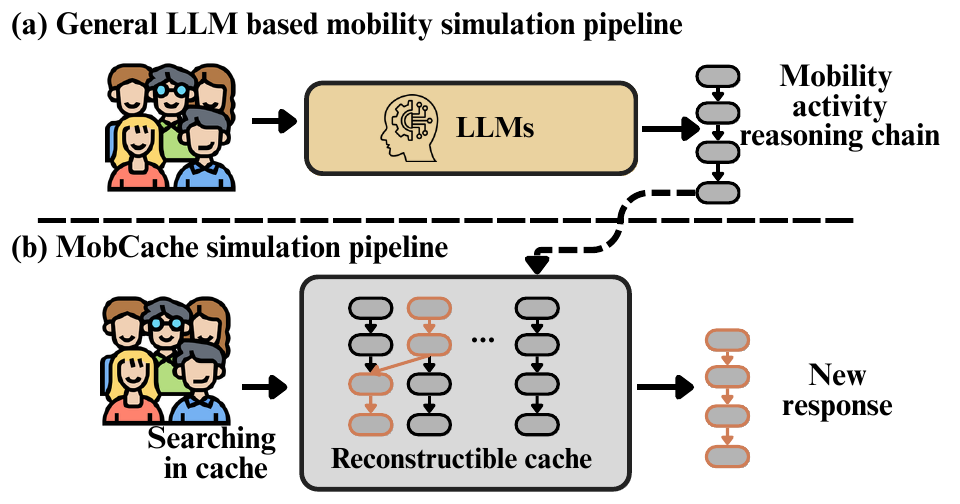}
    \vspace{-10pt}
    \centering
    \captionsetup{font={small}}
    \caption{Core idea of \name{}.}
    \label{fig:pipeline}
\vspace{-15pt}
\end{figure}

To overcome this limitation, we design a new caching paradigm: rather than caching an LLM's final responses, we cache the intermediate reasoning steps it produces while generating mobility data.
This design builds on prior work~\cite{shao2024chain,jiawei2024large} showing that mobility behaviors are typically produced through multi-step reasoning rather than a single step.
We represent these reasoning steps as nodes in a tree, so that cached steps can be recombined into new reasoning chains (Figure~\ref{fig:pipeline}).
This recombination lets a single cached step serve many responses, improving both reuse and diversity.
We call the resulting approach \textit{reconstructible caches}.
A straightforward implementation would keep each step as natural-language text and build new chains by concatenating steps from different cached chains.
This increases diversity, but it cannot ensure mobility awareness, such as the spatial and temporal constraints inherent to human movement~\cite{zhang2024spatial,wang2024spatiotemporal}, and therefore limits fidelity.

To this end, we design a mobility-aware cache framework that builds reconstructible caches to enable efficient large-scale human mobility simulations.
The framework makes mobility-aware reasoning chains reusable and recombinable by addressing two challenges, one in reasoning and one in decoding:

\noindent\textbf{(1) Reasoning perspective}: 
Reasoning chains are hard to make mobility-aware when each step is an explicit language token, since such tokens cannot flexibly encode mobility-specific constraints.
We therefore move the reasoning process into a latent space~\cite{hao2024training,tan2025think}, representing each step as an embedding rather than as language tokens.
This brings two benefits.
First, mobility constraints can be written directly into the embeddings during training.
Second, the embeddings implicitly capture diverse reasoning patterns and decision paths~\cite{hao2024training,ruan2025reasoning}, so that logically consistent reasoning chains can be explored efficiently during mobility-aware decoding (see below).
We also train a latent-space evaluator, guided by a fine-tuned LLM, to identify valid reasoning paths.

\noindent\textbf{(2) Decoding perspective}:
After obtaining the reconstructed reasoning chains in the latent space, we design a lightweight decoder trained through mobility law-constrained distillation. 
This decoder efficiently translates latent reasoning chains into natural language. 
This strategy avoids repeated calls to the original LLMs for decoding while achieving comparable performance, thereby ensuring efficiency and fidelity in large-scale simulations.

In particular, our main contributions are as follows.
\begin{itemize}[leftmargin=*, nosep]

\item We introduce building reconstructible caches in latent space to accelerate large-scale LLM-based mobility simulation, aiming to improve both simulation efficiency and the diversity of the generated mobility data.

\item We present \name{}, a mobility‑aware cache framework with two key components:
(1) a reconstructible cache that stores latent-space reasoning embeddings from a fine-tuned LLM and supports tree-structured search, so cached reasoning can be flexibly recombined; 
and (2) a lightweight decoder, trained via mobility law-constrained distillation, that converts latent-space reasoning chains into natural language while preserving spatial and temporal consistency.

\item  We conduct extensive experiments showing that \name{} outperforms all baselines on efficiency metrics while staying comparable on quality metrics. It reduces inference time by at least 83.61\%, increases tokens per second by 53.26\%, and lowers cost by 71.82\%. In a case study, applying \name{} to a state-of-the-art simulation baseline reduces inference time by 65.49\% and cost by 40.71\% with no loss in quality.
\end{itemize}

\section{Motivation}
\subsection{Why cache diversity matters}
\label{sec:cache_matter}

We run a simple experiment to show that a cache with limited diversity degrades simulation quality. From a real-world dataset (Section~\ref{sec:eva}), we randomly sample 10,000 trajectories as ground truth.
We then simulate trajectories with LLM-archetypes~\cite{chopra2024limits}, a common group-based method.
It groups people into clusters by profile attributes such as occupation and income, and keeps only a small set of real trajectories, 1,000 across all clusters.
To generate 10,000 new trajectories, we assign 10,000 new users to the most similar cluster by profile and samples one stored trajectory from that cluster.
When the store is small, many users receive the same trajectory, which limits diversity.

We compare the ground‑truth and simulated sets on two standard mobility signals. (1) \textbf{Stay duration.} For each trajectory, we compute the mean duration of all stops, then compare the two distributions. 
(2) \textbf{Location coverage.} We divide the study area into 1\,\text{km}$\times$1\,\text{km} grids, identify the 15 grids visited most often in the ground-truth set, and count how often the baseline visits those same grids. 
\vspace{-10pt}
\begin{figure}[h]\centering
\begin{minipage}[t]{0.46\linewidth}
    \includegraphics[width=\linewidth, keepaspectratio=true]{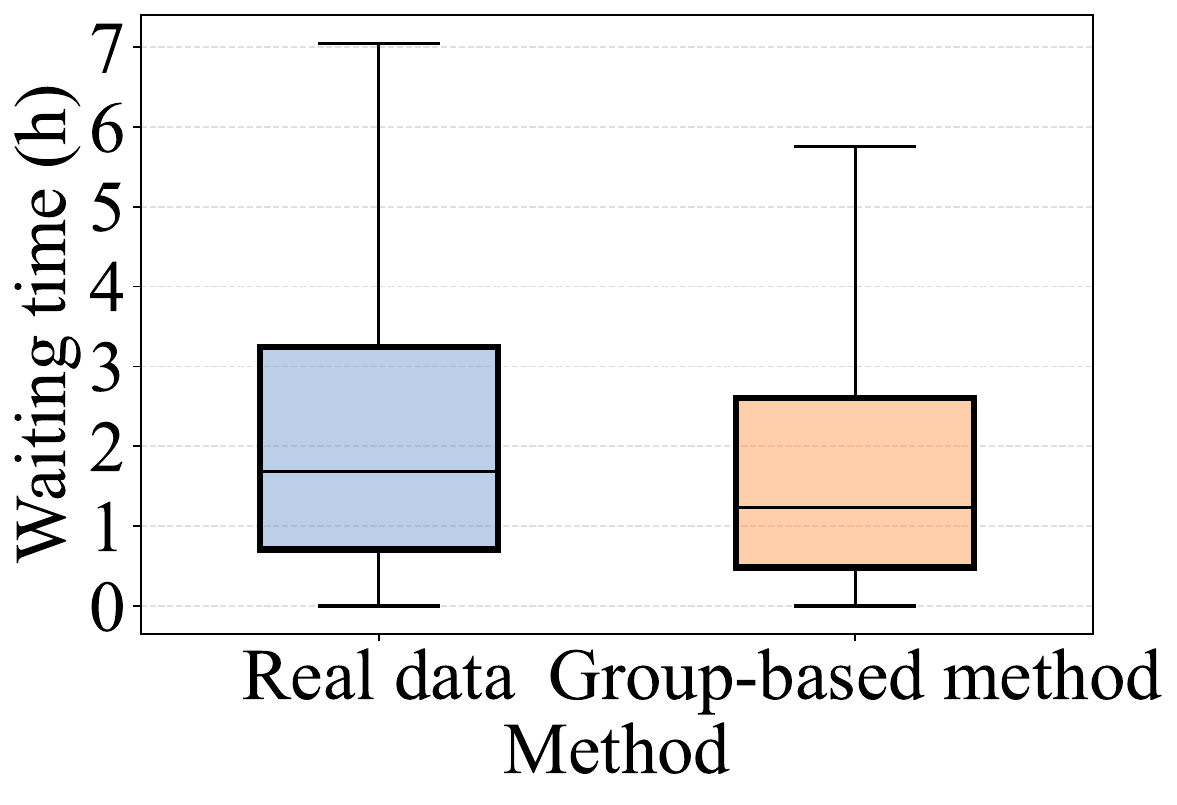}
    \vspace{-10pt}
    \captionsetup{
        font={small},
        skip=2pt  
    }
    \caption{Comparison of stay duration distributions.}
    \label{fig:duration}
\end{minipage}
\hspace{10pt}
\begin{minipage}[t]{0.46\linewidth}
    \includegraphics[width=\linewidth, keepaspectratio=true]{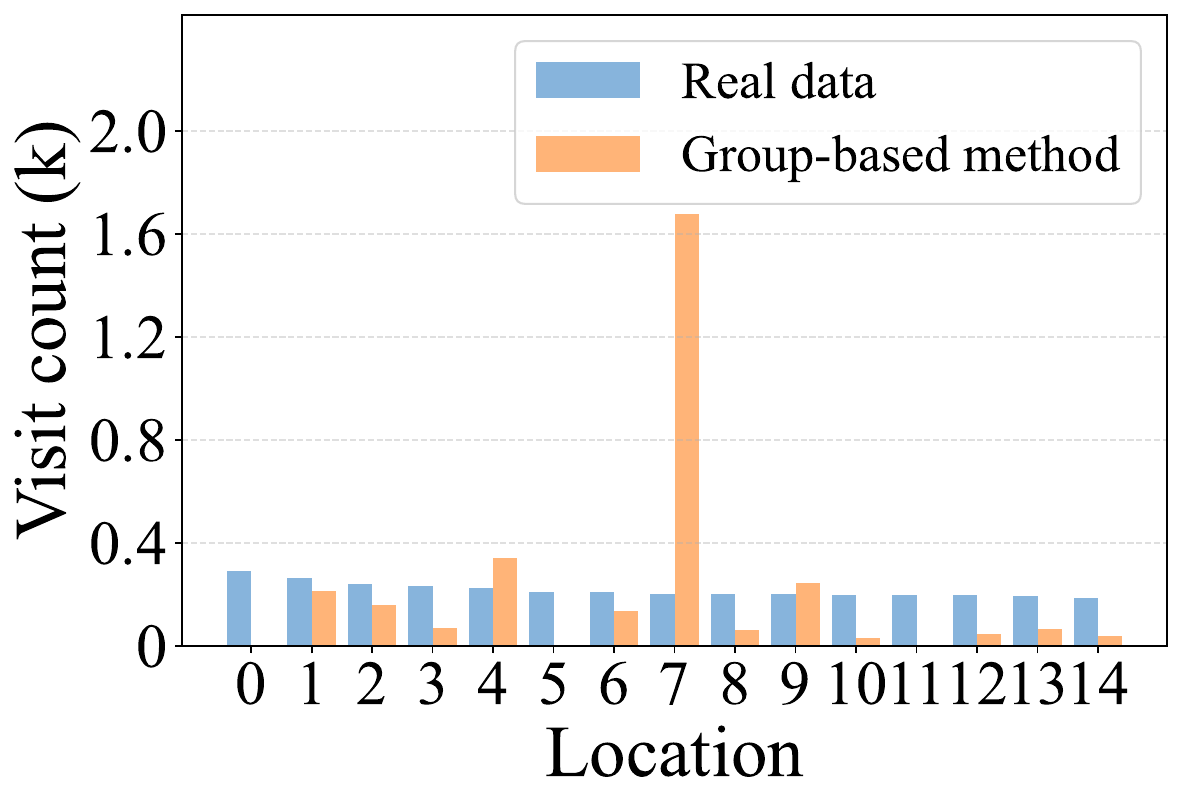}
    \vspace{-10pt}
    \captionsetup{
        font={small},
        skip=2pt  
    }
    \caption{Comparison of location coverage.}
    \label{fig:Location coverage}
\end{minipage}
\vspace{-12pt}
\end{figure}

Figure~\ref{fig:duration} compares the stay-duration distributions. Compared to the real data, the group-based method produces stay durations that are more tightly concentrated, indicating less behavioral diversity.
Figure~\ref{fig:Location coverage} illustrates the location coverage. The results show that the group-based approach underrepresents many of the most frequently visited locations in the real data.
Together, these results show that limited cache diversity reduces the realism of mobility simulation, which motivates a more flexible cache that can generalize beyond the stored samples.

\subsection{Logic inconsistency of reasoning chain reconstruction in language space}
A natural way to reuse cached reasoning chains is to recombine their steps directly in language space, that is, as natural-language text.
Suppose the cache holds many textual reasoning chains.
While following one chain, we may want to branch at some step and continue it with a plausible step drawn from another chain (as shown in Figure~\ref{fig:pipeline}).
A simple way to find such a step is to embed the current step and the candidate steps with a language model (e.g., \texttt{sentence-transformers/all-MiniLM-L6-v2}) and link two steps whenever their similarity exceeds a threshold, forming a new chain.

However, recombining steps this way has a key limitation: the text can read fluently while the underlying logic is wrong.
Box~\ref{box:bad-example} shows such a case.
The third step describes shopping for an \emph{upcoming holiday} on November 2nd, although no holiday falls near that date.
It also measures the shopping distance from the person's \emph{workplace}, yet the previous step left her at home and the day is a weekend, so she would not be at work.
The recombined steps are therefore logically inconsistent.

\begin{tcolorbox}[title=Bad example,label=box:bad-example, breakable]
This person is a 28-year-old woman. The education level is High school diploma, and consumption level is medium. 
Home and Workplace location: \texttt{(116.000, 40.000)}, ...;
Date: \texttt{2019-11-02}
\begin{itemize}[leftmargin=*, nosep]
    \item At 12:30 a.m., the woman returns home after a late-night social outing with friends, having spent the evening at a nearby restaurant. The distance traveled is approximately 2 km.
    \item At 10:00 a.m., she wakes up and decides to go for a morning walk in a nearby park to enjoy the fresh air. The park is about 1 km from her home.
    \item At 4:00 p.m., she heads to a shopping area to prepare for \textcolor{red}{\textbf{upcoming holiday events}}. The shopping area is about \textcolor{red}{\textbf{3 km from her workplace}}. 
\end{itemize}
\end{tcolorbox}
\vspace{-10pt}
\section{Design}
\subsection{Problem formulation}

Given individual profiles (age, occupation, and home locations) and the city’s points of interest (POIs), our goal is to design an efficient human mobility simulator. Taking the profiles and POIs as input, the simulator outputs a daily mobility trajectory for each individual. We represent a trajectory as a sequence of event-level spatiotemporal points rather than a continuously sampled GPS trace; each point is an activity event with a timestamp and latitude/longitude coordinates. The simulator should produce realistic trajectories while scaling to tens or hundreds of thousands of individuals.

\subsection{Overview}
\begin{figure*}[h]
\vspace{-10pt}
    \centering
    \includegraphics[width=0.92\linewidth, keepaspectratio=true]{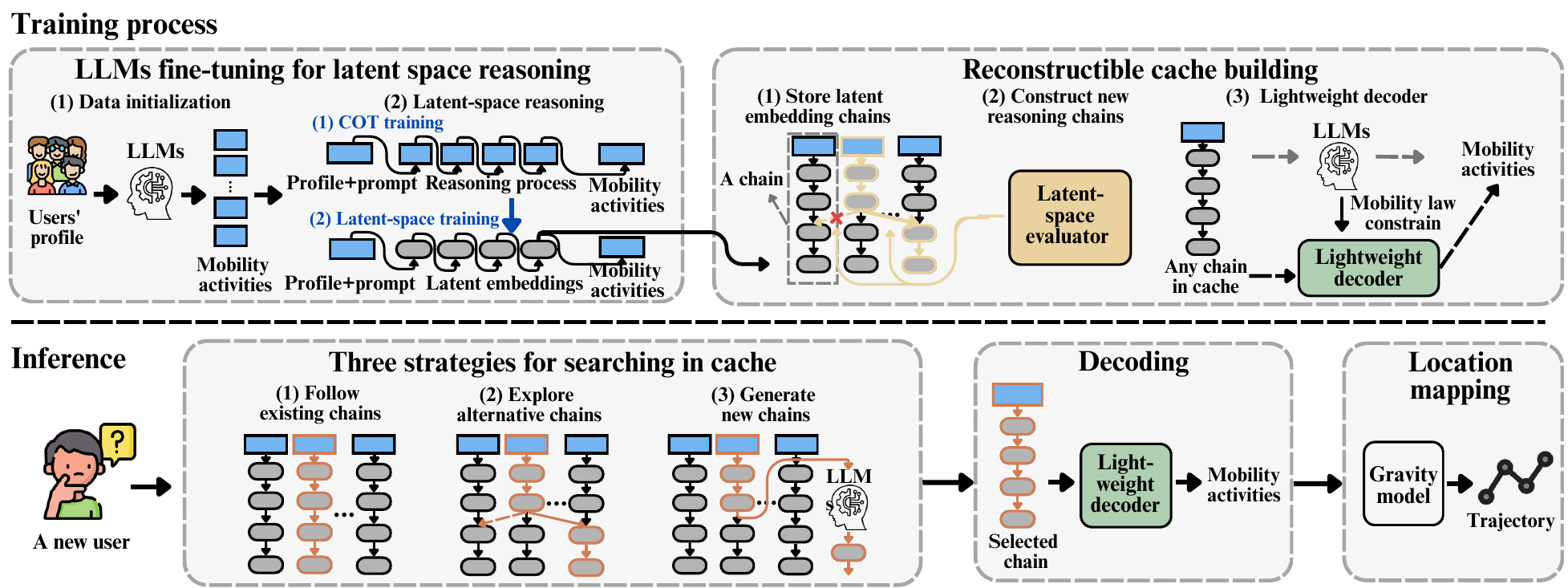}
    \vspace{-5pt}
    \captionsetup{font={small}}
    \caption{Framework of \name.}
    \label{fig:overview}
\vspace{-15pt}
\end{figure*}

We design \name{}, a mobility-aware caching framework that builds reconstructible caches for efficient, large-scale human mobility simulation.
Figure~\ref{fig:overview} shows its workflow, which has two phases: a \textbf{training phase} that builds reconstructible caches from LLM-generated trajectories, and an \textbf{inference phase} that generates new trajectories by reusing the caches.

\textbf{Training phase:}
We first use an existing LLM (e.g., GPT) with task-specific prompts to generate a small-scale mobility dataset; the same data can also be drawn from prior mobility simulators~\cite{shao2024chain}.
Each example pairs a \textbf{reasoning chain} with the resulting \textbf{mobility activities}, both in text, where the reasoning chain captures the intention behind each activity.
We then fine-tune the LLM on this data with a latent-space reasoning strategy~\cite{hao2024training} so that it reasons in latent space: each example now pairs a \textbf{latent-space reasoning chain} with the textual mobility activities.
We use these latent-space chains to build a reconstructible cache, whose construction has three parts:
(1) storage of latent-space reasoning chains, which supports tree-structured search for flexible reconstruction;
(2) a latent-space evaluator, which judges whether a new branch added to a chain is plausible; and
(3) a lightweight decoder, which turns latent-space reasoning chains back into textual mobility activities.

\textbf{Inference phase:}
For a new user, we first identify the cached users whose context (e.g., profile and date) is most similar.
Starting from a matched user, we obtain a latent-space reasoning chain for the new user, either by following an existing chain or by exploring the cache to build a new one.
We then decode the chain into textual mobility activities with the lightweight decoder and map those activities to real geographic locations with a mapping model.

\subsection{LLMs fine-tuning for latent-space reasoning}
\label{sec:llm_fine_tuning}

\subsubsection{\textbf{Data initialization}}
We first leverage general LLM-based human mobility simulation methods to generate a small-scale mobility dataset for cache construction. 
In our work, we use GPT-4o-mini as the generator with task-specific prompts, but this step can be replaced by other LLM-API-based human mobility simulation method~\cite{shao2024chain,bhandari2024urban}. 
Following prior work, each generated example is a text pair of a reasoning chain and its mobility activities, where the reasoning chain captures the intention behind each activity.
To ensure this format, we construct prompts that include the person’s demographic profile (e.g., income and occupation), the date, nearby POIs around their home and workplace, and other task-specific requirements. 
An example is provided in the appendix.

\subsubsection{\textbf{Latent-space reasoning}}
To make reasoning chains reusable, we move the reasoning process into the latent space~\cite{hao2024training,tan2025think}, where each reasoning step is represented as a latent-space embedding instead of a language token.
In standard chain-of-thought (CoT) reasoning training, the LLM is supervised to output both intermediate reasoning steps and the final answer as text.
In our latent-space reasoning training, we replaces the explicit reasoning step outputs with latent-space reasoning embeddings, following existing work~\cite{hao2024training}. 
Specifically, given an input prompt, the LLM is trained to produce latent-space reasoning embeddings $r_t$ at each reasoning step, which replace token-level outputs and are recursively fed back into the LLM as inputs for subsequent reasoning, while the final answer (i.e.,  the mobility activities) is still generated as text.
The process is formulated as
$r_{t} = f_\theta(q, R_{1:t-1}),$
where $q$ is the input prompt and $ R_{1:t-1} = [r_1, r_2, \ldots, r_{t-1}]$ are previous latent-space reasoning embeddings. The latent-space reasoning embedding $r_{t}$ is used as input for generating the next latent-space reasoning embedding.

\textbf{Multi-stage training process:}
We achieve the transition from CoT reasoning to latent-space reasoning through multi-stage training.
The LLM is first fine-tuned using standard CoT training.
We then replace the textual reasoning steps with latent embeddings one stage at a time, from the earliest step to the latest.
Specifically, at each stage $t$, the first $t$ reasoning steps are replaced by latent embeddings $R_{1:t}$, while the remaining reasoning steps and the final answer are still supervised in text.
This continues until all reasoning is performed in the latent space, while the final answer $y$ (i.e., mobility activities) remains under language-level supervision: 
$\max_\theta \ \log p_\theta(y \mid q, R_{1:t}).$

\subsection{Reconstructible cache building}

\subsubsection{\textbf{Store latent-space reasoning chains}}
Each cache entry stores one chain, consisting of (1) the \textbf{prompt} $q$, the initial task input; and (2) the \textbf{latent-space reasoning embeddings} $R_{1:t} = [r_1, r_2, \ldots, r_t]$, the intermediate reasoning steps, where $r_t$ is the embedding at step $t$.
Treating each embedding as a node, the cached chains together form a tree: a branch can extend from any intermediate node of one chain to nodes in other chains.
This shared structure is what enables tree-structured search and reasoning recombination.

\subsubsection{\textbf{Construct new reasoning chains}}
After storing the latent-space reasoning chains in the cache, we can further utilize these cached reasoning chains to generate new reasoning chains by branching from existing chains. 
However, latent-space reasoning embeddings are not interpretable by humans, making it hard to determine which branches are logically valid and consistent with human reasoning.

We therefore train a latent-space evaluator that scores how promising a branch is.
Formally, the latent-space evaluator model $g_\xi$ is trained to predict a score: $\sigma_t = g_\xi(q, R_{1:t-1}, r_t), $
where q is the prompt and $R_{1:t-1} = [r_1, r_2, \ldots, r_{t-1}]$ is the sequence of previous reasoning steps. $\sigma_t$ indicates the estimated quality of including $r_t$ as the next reasoning step. 
We implement the evaluator with a Transformer, which captures the contextual dependencies within a chain when judging coherence.

\textbf{Label construction:}
To construct labels for latent-space evaluator training, we adopt a similarity-based method:
\textbf{(1) Generate next latent-space reasoning step.} For each training example $(q, R_{1:t-1})$, we use the fine-tuned LLM (details in Section.~\ref{sec:llm_fine_tuning}) to generate the next latent-space reasoning steps: $\hat r_t$.
\textbf{(2) Compute similarity-based labels.} Given a candidate reasoning $r_t$, its supervision label $b_t$ is computed as: $b_t = \text{sim}(r_t, \hat r_t),$ 
    where $ \text{sim}(\cdot, \cdot)$ is a similarity function (e.g., embedding cosine similarity or model-based scoring). Intuitively, $b_t$ is high if $r_t$ is similar to next latent-space reasoning steps: $\hat r_t$, and low otherwise.
The evaluator is trained to minimize the squared error between the predicted scores and supervision labels $\mathcal{L} = \sum_t (\sigma_t - b_t)^2$.

\subsubsection{\textbf{Lightweight decoder}} 
Finally, we need to convert the cached latent-space reasoning chains into textual activities.
The LLM from Section~\ref{sec:llm_fine_tuning} can do this, but relying on it for every decode is expensive, so we train a small decoder to take its place.

\textbf{Distillation:}
Given a reasoning embedding chain $R = (r_1,\dots,r_T)$, the decoder generates an activity sequence $Y=(y_1,\dots,y_L)$ autoregressively, where $r_t$ is the embedding of the $t$-th reasoning step and $y_l$ is the $l$-th output token.
We treat the original LLM as the \textbf{teacher decoder} and train a \textbf{lightweight student decoder} to reproduce the same mapping from $R$ to $Y$, using cross-entropy loss:
$\mathcal{L}_{\text{distill}} = - \sum_{l=1}^{L} \log P_\rho\!\left(y_l \mid y_{<l}, \hat{R}\right), $
where $\rho$ are the student decoder's parameters, $y_l$ is the $l$-th token produced by the teacher, and $\hat{R}$ is the student-side input.
Because the teacher's embeddings $R$ and the student differ in dimensionality, we add an MLP projector $e_{\mu}$ that maps each teacher embedding into the student's space, giving $\hat{R}=e_{\mu}(R)$ as the decoder input.

\textbf{Mobility law constraint:}
To improve distillation quality, we add explicit mobility-law constraints.
The idea is to penalize differences between the mobility-law statistics of the student's outputs and the teacher's.
From the teacher's activity sequences $Y_{\text{teacher}}$, a statistics function $d(\cdot)$ extracts mobility-law features (e.g., jump distances), giving a target distribution $p_{\text{teacher}}(d(Y))$.
To keep the student side differentiable, we learn a function $z_\tau$ (e.g., an MLP) that predicts the mobility-law distribution from the student's hidden states $h_{\text{light}}$.
We then push $z_\tau(h_{\text{light}})$ toward the teacher distribution by minimizing the Kullback–Leibler (KL) divergence:
$\mathcal{L}_{\text{law}} = \mathrm{KL}\!\left( z_\tau(h_{\text{light}}) \ \|\ p_{\text{teacher}}(d(Y)) \right).$

\textbf{Training objective:}
The training objective combines the distillation loss $L_{\text{distill}}$ with the mobility law constraint loss $L_{\text{total}}=L_{\text{distill}}+\lambda L_{\text{law}},$
where $\lambda$ is a trade-off parameter.

\subsection{Inference}
To generate a trajectory for a new user, we first find the cached users whose context (e.g., profile and date) is most similar.
If no cached user passes a predefined similarity threshold, we call the fine-tuned LLM to generate a fresh reasoning chain and its activities, and store them in the cache for reuse.
If a similar cached user exists, we search the cache for a reasoning chain using one of the following strategies:

\begin{itemize}[leftmargin=*, nosep]
\item \textbf{Follow existing chains.}
We directly use the reasoning chain of the most similar cached user for the new user.

\item \textbf{Explore alternative chains.}
Starting from the reasoning chain of the retrieved similar user, we go through the chain and randomly select a node as a branching point. From this point, we search the cache for several candidate latent-space embedding nodes from other reasoning chains based on embedding similarity. 
A latent-space evaluator then scores these candidates and selects the most plausible one. 
The selected node is appended to the current chain, after which we repeat the above procedure by selecting a new branching point and extending the chain again. 
This process is repeated for a fixed number of iterations.

\item \textbf{Generate new chains.}
If the latent-space evaluator assigns low scores to all available branching nodes, we invoke the LLM to generate a new reasoning chain, which is then stored in the cache for future reuse.
\end{itemize}
We use an exploration rate to control the probability that the model follows an existing chain or explores a new branch. 
After obtaining a valid reasoning chain using these strategies, we input it into the lightweight decoder to decode the corresponding textual mobility activities. Finally, these textual activities are mapped to real geographic locations to form trajectories using the gravity model.

\section{Evaluation}
\label{sec:eva}

\begin{table*}[t]
\small
\centering
\renewcommand{\arraystretch}{1}
\setlength{\tabcolsep}{3pt}
\caption{Overall performance on the Beijing and NYC check-in datasets. Bold indicates the best result, and underline indicates the second-best result. Columns marked with $\downarrow$ indicate that lower values are better, while columns with $\uparrow$ indicate that higher values are better.}
\vspace{-10pt}
\label{tab:overall}
\resizebox{\textwidth}{!}{%
\begin{tabular}{llccc|ccccc}
\toprule
\multirow{2}{*}{Dataset} & \multirow{2}{*}{Method} & \multicolumn{3}{c|}{\textbf{Efficiency}} & \multicolumn{5}{c}{\textbf{Quality}} \\
\cmidrule(lr){3-5} \cmidrule(lr){6-10}
 & & Inference time $\downarrow$ & Tokens $\uparrow$ & Cost (1e-2) $\downarrow$ & Radius $\downarrow$ & Duration $\downarrow$ & Distance $\downarrow$ & Locfreq $\downarrow$ & Odsim $\downarrow$ \\
\midrule
\multirow{5}{*}{Beijing}
 & CoPB    & 45.8700$\pm$2.7767 & 40.4406$\pm$1.9513 & 0.6129$\pm$0.0045 & 0.0863$\pm$0.0037 & 0.0314$\pm$0.0005 & 0.0315$\pm$0.0004 & 0.0216$\pm$0.0002 & 0.3213$\pm$0.0013 \\
 & UML     & \underline{8.6500$\pm$0.2121} & 71.4088$\pm$1.5432 & \underline{0.0700$\pm$0.0017} & \underline{0.0823$\pm$0.0141} & \textbf{0.0194$\pm$0.0004} & 0.0569$\pm$0.0051 & \underline{0.0191$\pm$0.0016} & \underline{0.2887$\pm$0.0029} \\
 & CitySim & 35.2400$\pm$1.7536 & \underline{74.5850$\pm$1.0536} & 0.2445$\pm$0.0006 & 0.0746$\pm$0.0139 & 0.0252$\pm$0.0004 & 0.0287$\pm$0.0058 & 0.0328$\pm$0.0024 & 0.3445$\pm$0.0151 \\
 & LLMob   & 18.7467$\pm$0.7679 & 54.8446$\pm$2.1229 & 0.2287$\pm$0.0022 & 0.0986$\pm$0.0097 & 0.0322$\pm$0.0003 & \underline{0.0298$\pm$0.0055} & 0.0382$\pm$0.0015 & 0.3606$\pm$0.0081 \\
\cmidrule(lr){2-10}
 & \name{} & \textbf{1.2723$\pm$0.0232} & \textbf{119.4150$\pm$3.3305} & \textbf{0.0177$\pm$0.0003} & \textbf{0.0592$\pm$0.0033} & \underline{0.0218$\pm$0.0002} & \textbf{0.0271$\pm$0.0048} & \textbf{0.0189$\pm$0.0036} & \textbf{0.2634$\pm$0.0108} \\
\midrule
\multirow{5}{*}{NYC}
 & CoPB    & 38.6933$\pm$1.4975 & 46.9200$\pm$1.7446 & 0.6222$\pm$0.0007 & 0.1675$\pm$0.0063 & N/A & 0.0524$\pm$0.0058 & 0.0286$\pm$0.0024 & 0.1784$\pm$0.0093 \\
 & UML     & \underline{8.7000$\pm$0.1414} & \underline{70.9510$\pm$3.2031} & \underline{0.0703$\pm$0.0023} & 0.1665$\pm$0.0178 & N/A & \underline{0.0260$\pm$0.0057} & 0.0236$\pm$0.0023 & \underline{0.1517$\pm$0.0204} \\
 & CitySim & 37.8250$\pm$1.2233 & 68.3798$\pm$1.4564 & 0.2661$\pm$0.0018 & \underline{0.1240$\pm$0.0281} & N/A & 0.0685$\pm$0.0092 & \underline{0.0231$\pm$0.0029} & 0.2325$\pm$0.0131 \\
 & LLMob   & 15.3300$\pm$0.6180 & 57.5600$\pm$2.0219 & 0.3474$\pm$0.0225 & 0.1331$\pm$0.0052 & N/A & 0.0425$\pm$0.0058 & 0.0855$\pm$0.0021 & 0.2167$\pm$0.0115 \\
\cmidrule(lr){2-10}
 & \name{} & \textbf{1.4259$\pm$0.0239} & \textbf{108.7361$\pm$1.1764} & \textbf{0.0198$\pm$0.0003} & \textbf{0.1082$\pm$0.0146} & N/A & \textbf{0.0206$\pm$0.0016} & \textbf{0.0216$\pm$0.0003} & \textbf{0.1485$\pm$0.0168} \\
\bottomrule
\end{tabular}%
}
\vspace{-5pt}
\end{table*}

\subsection{Dataset description}
We use two public mobility datasets. 
One dataset was collected in Beijing, China, covering the period from October 1, 2019 to December 31, 2019, and includes mobility records for 100,000 individuals~\cite{shao2024chain}. 
The dataset contains mobility trajectories and user profile information (e.g., age, gender, and occupation), collected via a social networking platform.
Another dataset is the NYC POI check-in dataset~\cite{yang2014modeling}. For this dataset, we simulate users’ profiles based on U.S. Census demographics~\cite{census}.

\subsection{Evaluation setup}
\subsubsection{\textbf{Evaluation configuration}}
For the Beijing dataset, we preprocess the raw mobility records by removing users with incomplete profiles and extracting stay points using a 20-minute, 500-meter threshold following prior work~\cite{gonzalez2008understanding,song2010modelling}. 
From the preprocessed dataset, we select approximately 1,000 users and use an LLM to generate 13,000 synthetic human mobility trajectories based on their profiles.
We fine-tune an LLM on these synthetic trajectories (Section~\ref{sec:llm_fine_tuning}) so that it reasons in latent space, and store the resulting latent-space reasoning chains as our mobility cache.
For evaluation, we sample a separate set of 20,000 real trajectories as the test set, with no user overlap with the cache, and compare the generated trajectories against it.

For the NYC POI check-in dataset, we select approximately 400 users and use an LLM to generate two weeks of synthetic mobility trajectories based on their profiles. We fine-tune an LLM on these synthetic trajectories for mobility cache building (Section~\ref{sec:llm_fine_tuning}).
For evaluation, we construct a separate test set from six months of real check-in trajectories, containing approximately 16,000 trajectories.

\subsubsection{\textbf{Implementation}}
We implement our method using Python 3.10 and PyTorch 2.1.0.
\textbf{(1) Latent-space reasoning.} we fine-tune a LLaMA-3.2-3B model using a single NVIDIA A100 GPU. Following the procedure in~\cite{hao2024training}, we use 6-stage training (stages 0--5) with a batch size of 7. In stage 0, the model is fine-tuned with standard token-level autoregressive training for 2 epochs; in each later stage $k$ ($k=1,\dots,5$), the first $k$ reasoning steps are replaced by $k$ latent embeddings and the model is trained for 1 epoch.
\textbf{(2) Decoder training.} we distill a LLaMA-3.2-1B decoder from a teacher LLaMA-3.2-3B on a single NVIDIA A100 GPU, using the same 6-stage training (1 epoch per stage) with batch size 7.
\textbf{(3) Cache inference.} to determine profile similarity during cache retrieval, we employ the pre-trained language model~\cite{reimers-gurevych-2019-sentence}. 
We test the exploration rate in [0.3, 0.5, 0.7] and select 0.5 based on the efficiency and quality trade-off. 
We test $\lambda$ in [0.01, 0.03, 0.05, 0.07] and select 0.05 based on the best performance.
The number of search rounds is sampled uniformly between 1 and 3.

\subsubsection{\textbf{Baselines}}
We compare our method with LLM-based human mobility simulation baselines including:

\begin{itemize}[leftmargin=*, nosep]
    \item \textbf{CoPB~\cite{shao2024chain}} is a mobility simulation framework guides LLMs through reasoning stages to generate mobility activities.

    \item \textbf{Urban-Mobility-LLM (UML)
~\cite{bhandari2024urban}} is a method that synthesizes travel survey data by prompting LLMs to generate individual mobility patterns.

    \item \textbf{LLMob~\cite{jiawei2024large}} uses LLM-based agents for personal mobility simulation, enhanced by self-consistency and retrieval strategies.

    \item \textbf{CitySim~\cite{bougie2025citysim}} is an LLM-driven urban agent simulation framework that generates daily activity through planning, memory, needs, and spatial decision-making.    
\end{itemize}
We provide additional details on the adaptation and implementation of these baselines in Appendix~\ref{app:baselines}.

We also compare three variants of our model: (1) \textbf{w/o LE} removes the latent-space evaluator and selects branches with a plain similarity function instead; (2) \textbf{w/o MD} removes the mobility-law distillation loss; and (3) \textbf{w/o LD} removes the lightweight decoder and decodes latent embeddings with the original LLM.

\subsubsection{\textbf{Metric}}
We evaluate our model from two axes: efficiency and simulation quality.

\textbf{Efficiency evaluation:} We measure efficiency in three ways.
(1) \emph{Inference time} (s/trajectory): the average inference time required to generate a single simulated trajectory.
For a fair comparison, every method runs sequentially, without parallel or batched execution.
(2) \emph{Tokens} (tokens/s): the number of output tokens generated per second during inference.
(3) \emph{Monetary cost} (\$/trajectory): the average cost per trajectory when generating 20K trajectories. For the API-based baselines (UML, CoPB, LLMob, CitySim), we estimate the average input and output tokens per trajectory and apply GPT-4o-mini's published rates of \$0.15 per 1M input tokens and \$0.60 per 1M output tokens. For our method, which runs a local LLM, we multiply the average inference time per trajectory by a GPU rate of \$0.50/hour on an A6000.

\textbf{Quality evaluation:} To evaluate the similarity between generated and real-world mobility data, we compute the Jensen–Shannon divergence (JSD) between their distributions on five key metrics following existing work~\cite{shao2024chain,gonzalez2008understanding,song2010modelling}: (1) \emph{Radius of gyration} measures the spatial dispersion of individual mobility. We compute the radius of gyration for each individual as the root mean square distance from all visited points to the individual’s trajectory centroid.
(2) \emph{Stay duration} measures how long individuals remain at each visited location. We compute the duration distribution from the time intervals between consecutive movements.
(3) \emph{Distance} measures the distance between consecutive visited locations. We calculate it as the geographical distance between two consecutive locations in a trajectory. 
(4) \emph{Location frequency} (Locfreq) measures how well the generated trajectories replicate the spatial distribution of visits. We compute location frequency as the JSD between the spatial visit frequency distributions of the generated and real trajectories.
(5) \emph{Origin-destination similarity} (Odsim) measures the similarity between generated and real trajectories in terms of OD travel patterns. It is computed by comparing the normalized frequency distributions of OD pairs using JSD.

\subsection{Overall performance}
\label{sec:overall}
\textbf{RQ1:} \textit{How does \name{} perform compared with existing LLM-based mobility simulation baselines in terms of efficiency and simulation quality?}
\subsubsection{\textbf{Performance on Beijing dataset}}
Table~\ref{tab:overall} reports the comparison between our method and various baselines on the Beijing dataset.
\textbf{In terms of efficiency}, our method outperforms all the baselines across multiple metrics,
achieving at least a 85.29\% improvement in inference time, a 60.10\% improvement in tokens/s, and a 74.71\% reduction in cost.
It is worth mentioning that CoPB has a longer inference time mainly because it adopts an explicit multi-stage reasoning process, where each time step sequentially performs intention-confidence ranking, activity selection, and duration estimation while recursively conditioning on previously generated activities.
Regarding simulation quality, our method achieves comparable results to LLM-based approaches, demonstrating the effectiveness of our method.

\subsubsection{\textbf{Performance on NYC check-in dataset}}
\label{sec:nyc}
We also conduct experiments on the NYC POI check-in dataset~\cite{yang2014modeling}. Specifically, we simulate users' profiles based on U.S. Census demographics~\cite{census}, and use these profiles as inputs to generate human mobility trajectories.
The NYC results use slightly different evaluation metrics because this dataset is based on check-in records, where intervals between check-ins may not accurately reflect actual stay durations. Therefore, duration is not compared on the NYC dataset.
The results are shown in Table.~\ref{tab:overall}. In terms of efficiency, our method outperforms all baselines across multiple metrics, achieving at least a 83.61\% reduction in inference time, an 53.26\% increase in tokens/s, and a 71.83\% improvement in cost efficiency.
In terms of simulation quality, our method achieves performance comparable to LLM-based approaches, demonstrating the effectiveness of our framework.

\vspace{-5pt}
\subsection{Evaluation of decoding results via LLMs}
\textbf{RQ2:} \textit{Can the decoded trajectories preserve profile consistency and mobility behavioral coherence?}
Beyond quantitative metrics, we further assess decoded trajectories from two further angles: (1) profile consistency, whether the generated mobility activities align with the user's profile, and (2) mobility behavioral coherence, whether the generated sequence remains temporally, spatially, and semantically consistent after latent-space decoding.

To evaluate profile consistency, we employ an external LLM evaluator~\cite{liu2023g, zheng2023judging}. For each sample, the evaluator is provided with the user profile and the decoded mobility activity sequence, and is asked to determine whether the sequence is plausible for that individual. We then compute the proportion of sequences receiving a positive judgment. The results show that 91\% of the trajectories generated by our method are considered consistent with the associated user profiles, compared with 82\% for \name{} w/o LE. This improvement suggests that the latent-space evaluator effectively guides cache-path selection and helps preserve profile-aware behavioral patterns during decoding.

To evaluate semantic consistency under latent recombination, we add an experiment on behavioral coherence. We compare \name{} with a random-recombination baseline that merges latent embeddings without any compatibility check. An independent LLM evaluator judges whether each generated trajectory is coherent as a whole, temporally, spatially, and semantically. Our method achieves a coherence rate of 87.5\%, substantially higher than the 52.0\% obtained by random recombination. These results indicate that the proposed latent-space evaluation mechanism preserves meaningful behavioral structure and mitigates the semantic inconsistencies that may arise from unconstrained latent recombination.

\subsection{Importance of cache diversity}
\label{sec:eva_group}
\textbf{RQ3:} \textit{How important is cache diversity for preserving fidelity and behavioral diversity in mobility simulation?}
In Section~\ref{sec:cache_matter}, we argue that cache diversity is essential for realistic mobility simulation. We test this by comparing \name{} with the group-based approach~\cite{chopra2024limits} on the Beijing dataset, from two angles: population-level fidelity and population-level behavioral diversity.

\begin{figure}[h]
    \centering
    \includegraphics[width=\linewidth, keepaspectratio=true]{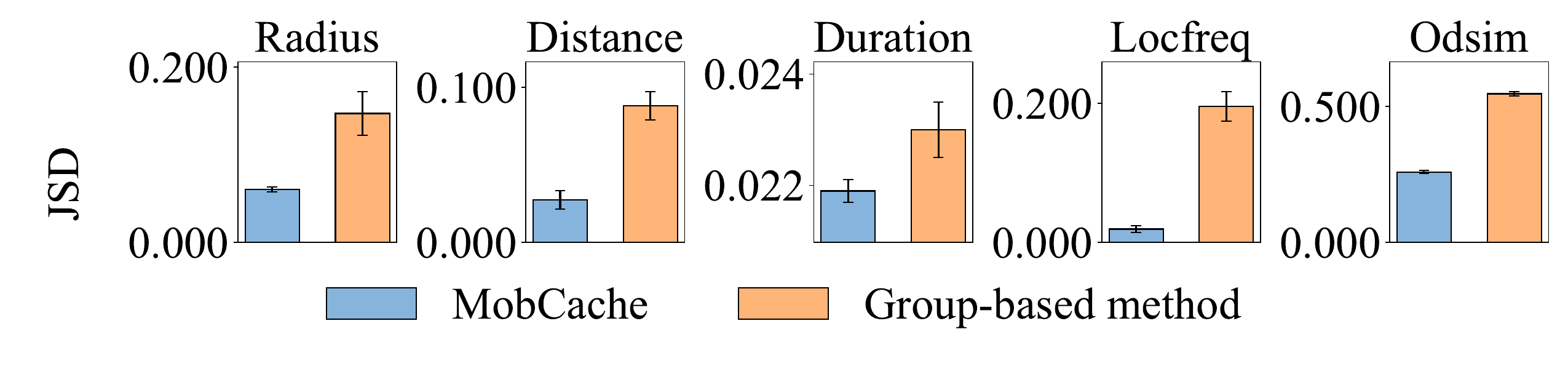}
    \vspace{-15pt}
    \captionsetup{font={small}}
    \caption{Effect of cache diversity on simulation fidelity.}
    \label{fig:group_fidelity}
\vspace{-10pt}
\end{figure}

First, cache diversity improves the fidelity of simulation. Figure~\ref{fig:group_fidelity} reports the JSD between simulated and real distributions for both methods across the five metrics. Averaged over the five, \name{} improves JSD by 55.27\% relative to the group-based method, showing that a diverse cache reproduces real mobility statistics more faithfully.

\begin{figure}[h]\centering
\vspace{-5pt}
\begin{minipage}[h]{0.46\linewidth}
    \centering
    \includegraphics[width=\linewidth, keepaspectratio=true]{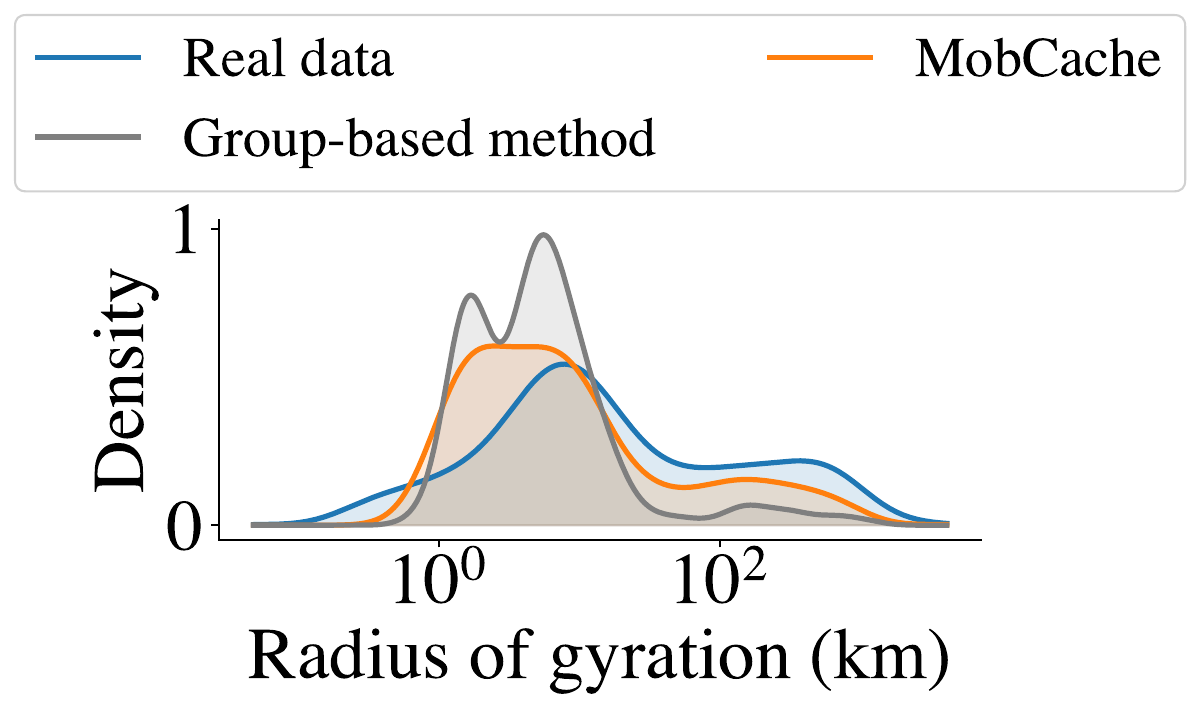}
    \vspace{-15pt}
    \captionsetup{font={small}}
    \caption{Effect of cache diversity on population diversity.}
    \label{fig:group_diversity}
\end{minipage}
\hspace{2pt}
\begin{minipage}[h]{0.46\linewidth}
    \includegraphics[width=\linewidth, keepaspectratio=true]{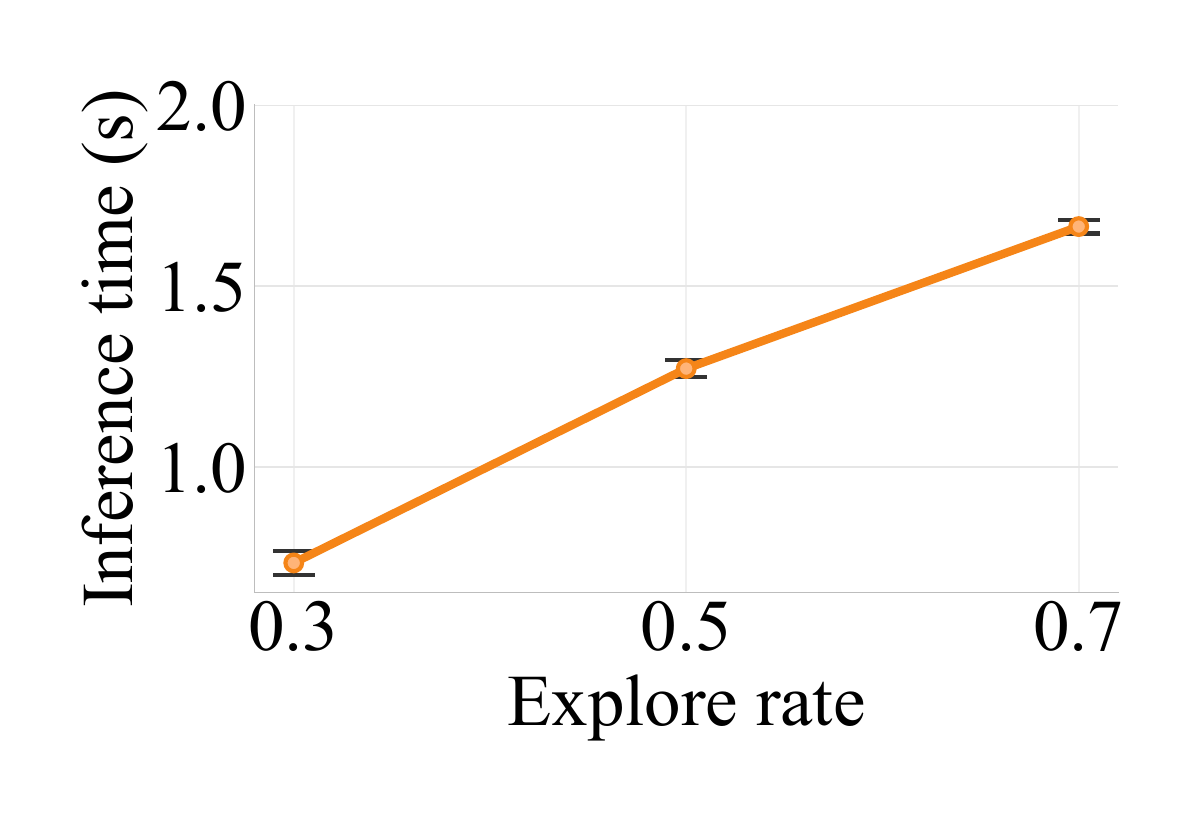}
    \vspace{-15pt}
    \captionsetup{
        font={small},
        skip=2pt  
    }
    \caption{Impact of exploration rate on inference time.}
    \label{fig:explore_rate_time}
\end{minipage}
\vspace{-10pt}
\end{figure}

Second, cache diversity is crucial for preserving behavioral heterogeneity. Figure~\ref{fig:group_diversity} shows the radius-of-gyration distribution for the real and simulated populations.
\name{} tracks the real distribution closely, whereas the group-based method underestimates the highly mobile individuals in the tail and so spans a much narrower range of behaviors.
A handful of archetypes cannot capture the full diversity of real movement; by caching and recombining mobility-aware reasoning chains, \name{} generates more varied patterns and better preserves population heterogeneity.

\subsection{Cache reuse and exploration analysis}
\textbf{RQ4:} \textit{How does \name{} balance cache reuse and exploration to achieve efficient mobility simulation?}
To understand where the efficiency gains come from, we analyze the role of cache reuse and exploration in our framework. Unlike conventional caching systems, where efficiency is typically characterized by cache hit and miss rates, our method operates under a retrieval-based reuse mechanism. Specifically, for 10K users outside the cache, every user can consistently find a highly similar cached profile, with similarity scores exceeding 0.977. As a result, strict hit/miss distinctions become less informative, since suitable cache candidates are almost always available.

\begin{figure}[h]
\vspace{-10pt}
    \centering
    \includegraphics[width=\linewidth, keepaspectratio=true]{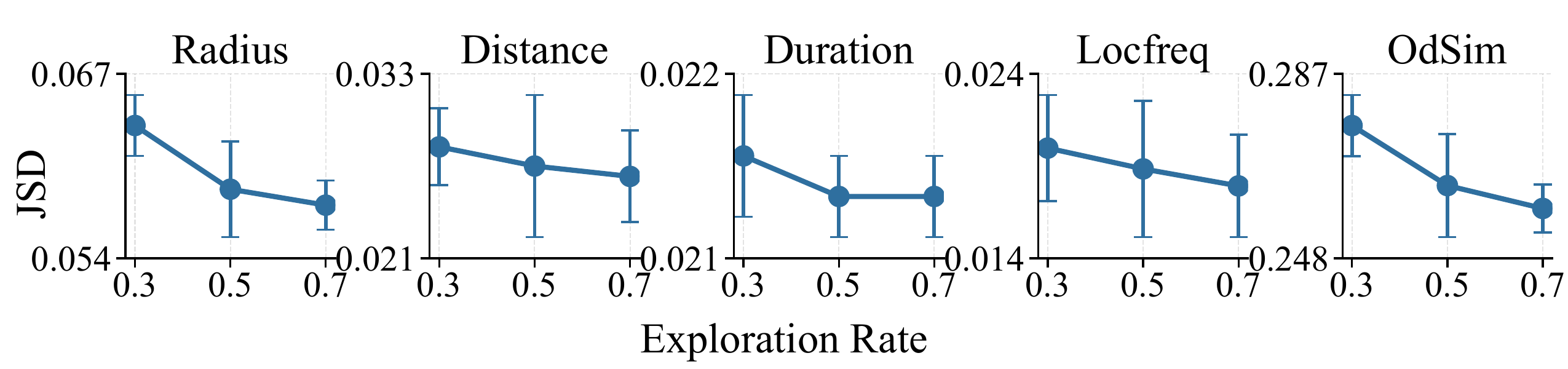}
    \vspace{-15pt}
    \captionsetup{font={small}}
    \caption{Impact of exploration rate on simulation quality.}
    \label{fig:exploration_quality}
\vspace{-15pt}
\end{figure}

Instead, the key factor governing efficiency is the exploration rate, which sets how often the system searches for new mobility chains beyond the retrieved ones.
We evaluate this effect on the Beijing dataset by varying the exploration rate and measuring both simulation quality and inference time. Figure~\ref{fig:explore_rate_time} shows that inference time increases with the exploration rate, as additional mobility chains must be searched and evaluated during simulation. Reducing the exploration rate from 0.7 to 0.3 more than halves the inference time, highlighting exploration as the primary source of online computational overhead.
However, lower exploration also reduces simulation quality. As shown in Figure~\ref{fig:exploration_quality}, increasing the exploration rate improves simulation quality, but the marginal gains beyond 0.5 are limited relative to the additional computational cost. Therefore, we use an exploration rate of 0.5 in all experiments.

\begin{figure*}[t]
    \centering
    \includegraphics[width=0.8\linewidth, keepaspectratio=true]{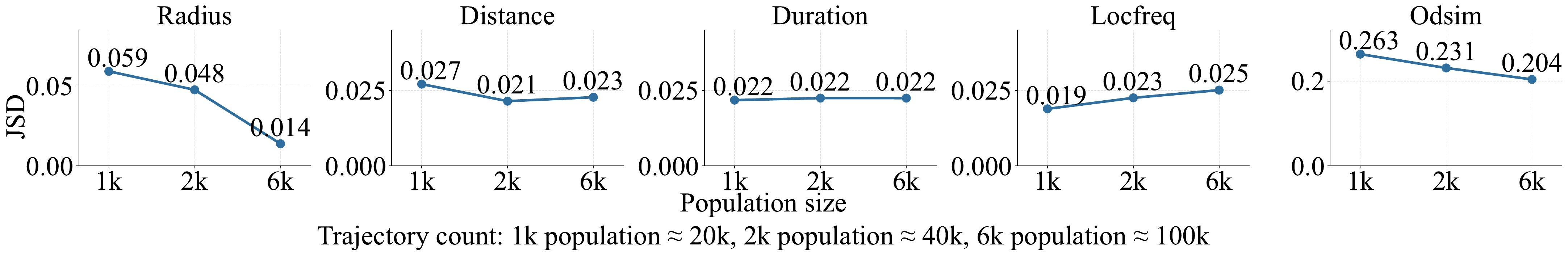}
    \vspace{-10pt}
    \captionsetup{font={small}}
    \caption{Scaling simulation by population size.}
    \label{fig:scaling_population}
\end{figure*}

\begin{figure*}[t]
\vspace{-10pt}
    \centering
    \includegraphics[width=0.8\linewidth, keepaspectratio=true]{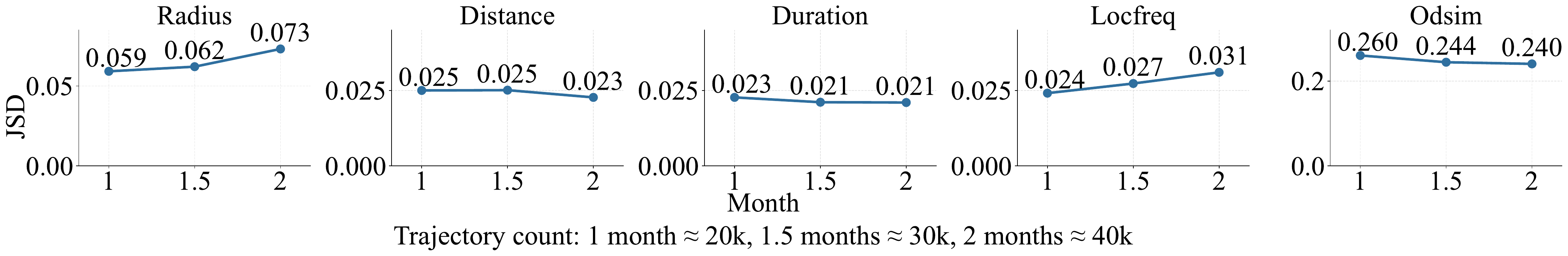}
    \vspace{-10pt}
    \captionsetup{font={small}}
    \caption{Scaling simulation by temporal horizon.}
    \label{fig:scaling_temporal}
\vspace{-15pt}
\end{figure*}

\subsection{Scalability under large-scale simulation}
\textbf{RQ5:} \textit{How well does \name{} scale under larger population sizes and longer simulation horizons?}
To further evaluate the scalability of our approach, we conduct two sets of extended experiments on the Beijing dataset by increasing the simulation scale along two dimensions: population size and temporal horizon. In both experiments, the cache is constructed from the original dataset and reused directly without any expansion or retraining, allowing us to assess whether the cached mobility patterns remain effective under substantially larger simulation scales.

\textbf{Scaling by population size.} We first increase the number of users from 1K to 6K, which increases the number of generated trajectories from approximately 20K to 100K. As shown in Figure~\ref{fig:scaling_population}, the performance remains stable despite the substantial increase in simulation scale. For most metrics, the JSD values exhibit only minor fluctuations. Notably, Radius and Odsim show improved alignment with the real data, with their JSD values decreasing from 0.059 to 0.014 and from 0.263 to 0.204, respectively. We attribute these improvements to the larger trajectory collection, which provides more reliable estimates of the underlying mobility distributions and reduces statistical variance. Overall, the results suggest that the cached mobility patterns generalize well to substantially larger populations and that our approach can support large-scale deployment without requiring cache expansion or retraining.

\textbf{Scaling by temporal horizon.} We further evaluate scalability with respect to the simulation duration. Specifically, we extend the temporal horizon from one month to two months, increasing the number of generated trajectories from approximately 20K to 40K while keeping the cache fixed. As shown in Figure~\ref{fig:scaling_temporal}, the performance remains largely stable under the extended simulation horizon. Across all mobility statistics, the JSD values exhibit only minor variations despite the doubled simulation period. These results demonstrate the robustness of our approach to temporal scaling without requiring cache expansion or retraining.

\subsection{Comparison with fine-tuning-based LLM mobility simulation}\label{sec:eval-fine-tune-llm}
\textbf{RQ6:} \textit{Can \name{} achieve competitive performance against fine-tuned LLM mobility simulation that uses real-world trajectories?}
We compare \name{} with Geo-LLaMA~\cite{li2024geo}, a representative fine-tuning-based LLM mobility simulation that relies on real trajectories for training. This comparison is supplementary to our main setting, where \name{} does not use any real trajectories from the target task for training.
Following its training-based setting, we fine-tune Geo-LLaMA using 7,000 real-world trajectories from the Beijing dataset. To avoid data leakage, the individuals included in the fine-tuning set are strictly disjoint from those in the test set.
Geo-LLaMA inference was performed locally on an NVIDIA RTX A6000 GPU using its LLaMA-2-7B-based implementation.

\begin{figure}[h]\centering
\begin{minipage}[h]{0.47\linewidth}
\vspace{-5pt}
    \includegraphics[width=\linewidth, keepaspectratio=true]{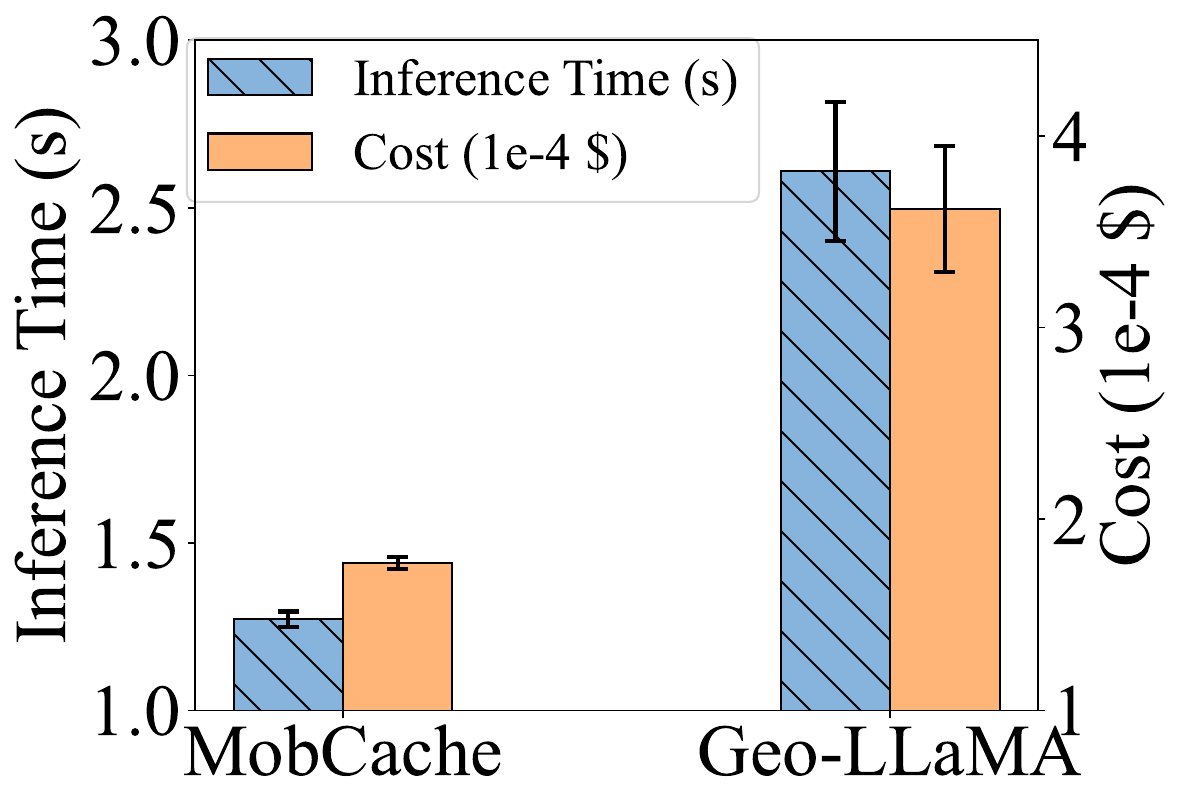}
    \vspace{-10pt}
    \captionsetup{
    font={small},
    skip=2pt  
}
    \caption{Efficiency comparison (\name{} vs. Geo-LLaMA).}
    \label{fig:geo_efficiency}
\end{minipage}
\hspace{5pt}
\begin{minipage}[h]{0.47\linewidth}
\vspace{-5pt}
    \includegraphics[width=\linewidth, keepaspectratio=true]{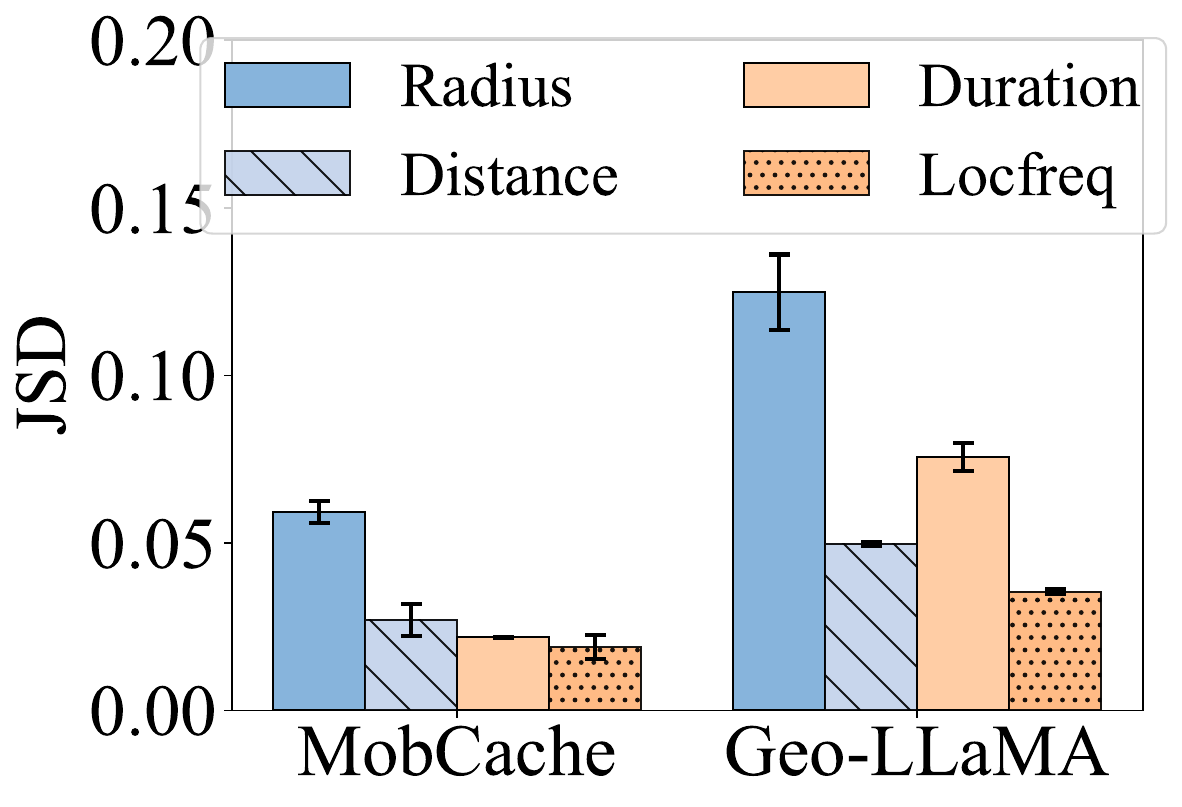}
    \vspace{-10pt}
    \captionsetup{
    font={small},
    skip=2pt  
}
    \caption{Quality comparison (\name{} vs. Geo-LLaMA).}
    \label{fig:geo_performance}
\end{minipage}
\vspace{-10pt}
\end{figure}

Figure~\ref{fig:geo_efficiency} and Figure~\ref{fig:geo_performance} shows that MobCache improves both efficiency and simulation quality over Geo-LLaMA. In terms of efficiency, MobCache achieves lower inference time and cost. In terms of simulation quality, MobCache obtains lower JSD across metrics. Although Geo-LLaMA is trained on real trajectories, its training and test users are disjoint; therefore, its performance still depends on how well it generalizes to unseen users. Under this setting, MobCache remains more effective despite not using real trajectories for training.

\subsection{Ablation study}
\textbf{RQ7:} \textit{How does each key component affect the performance of \name{}?}
\textbf{Effect of latent-space evaluator:}
To investigate the role of the latent-space evaluator, we substitute it with a naive similarity function (i.e., \name{} w/o LE) for determining the validity of connections between latent embeddings. 
The results (Beijing dataset) in Table~\ref{tab:ablation} show a decline in simulation quality, highlighting the evaluator’s critical role in ensuring coherent behavior generation.

\textbf{Effect of decoder:}
To evaluate the effectiveness of our lightweight decoder, we use the original LLM in Section~\ref{sec:llm_fine_tuning} to decode the reasoning chain (i.e., \name{} w/o LD) on Beijing dataset. As shown in the Table~\ref{tab:ablation}, although it achieves better quality on all metrics, it comes at a high computational cost. This shows that our lightweight decoder provides a balance between quality and efficiency.
We also evaluate the role of the mobility law distillation (i.e., \name{} w/o MD). 
The results in Table~\ref{tab:ablation} reveal that removing mobility law constraints leads to a quality performance drop.

\begin{table*}[t]
\small 
\centering
\renewcommand{\arraystretch}{1}
\setlength{\tabcolsep}{3pt}
\caption{Comparison with Variants of \name{}. Bold scores are for the best values.}
\vspace{-10pt}
\label{tab:ablation}
\begin{tabular}{lccc|ccccc}
\toprule
Method             & \multicolumn{3}{c|}{\textbf{Effiency}}                                                                              & \multicolumn{5}{c}{\textbf{Quality}}                                                                                                      \\ 
                   & \multicolumn{1}{c}{Inference time $\downarrow$} & \multicolumn{1}{c}{Tokens $\uparrow$} & Cost (1e-2) $\downarrow$ & \multicolumn{1}{c}{Radius $\downarrow$} & \multicolumn{1}{c}{Duration $\downarrow$} & \multicolumn{1}{c}{Distance $\downarrow$} & \multicolumn{1}{c}{Locfreq $\downarrow$} & Odsim $\downarrow$ \\ \midrule
w/o LE               & \multicolumn{1}{c}{1.2710$\pm$0.0165}               & \multicolumn{1}{c}{118.8533$\pm$2.2022}               &   0.0177$\pm$0.0002   & \multicolumn{1}{c}{0.0619$\pm$0.0031}       & \multicolumn{1}{c}{0.0232$\pm$0.0002}         & \multicolumn{1}{c}{0.0286$\pm$0.0052}            & \multicolumn{1}{c}{0.0195$\pm$0.0029}        & 0.2747$\pm$0.0097       \\ 
w/o MD & \multicolumn{1}{c}{1.2777$\pm$0.0154}               & \multicolumn{1}{c}{119.1867$\pm$2.1794}                &   0.0177$\pm$0.0002   & \multicolumn{1}{c}{0.0616$\pm$0.0028}       & \multicolumn{1}{c}{0.0226$\pm$0.0001}         & \multicolumn{1}{c}{0.0279$\pm$0.0055}            & \multicolumn{1}{c}{0.0193$\pm$0.0033}        &0.2738$\pm$0.0099       \\ 
w/o LD         & \multicolumn{1}{c}{2.1950$\pm$0.0477}               & \multicolumn{1}{c}{75.8777$\pm$1.0337}              &   0.0309$\pm$0.0014   & \multicolumn{1}{c}{\textbf{0.0564$\pm$0.0026}}       & \multicolumn{1}{c}{\textbf{0.0205$\pm$0.0002}}         & \multicolumn{1}{c}{\textbf{0.0257$\pm$0.0045}}            & \multicolumn{1}{c}{\textbf{0.0170$\pm$0.0029}}        & \textbf{0.2374$\pm$0.0095}      \\  \midrule

\name{}           & \multicolumn{1}{c}{\textbf{1.2723$\pm$0.0232}}               & \multicolumn{1}{c}{\textbf{119.4150$\pm$3.3305}}                & \textbf{0.0177$\pm$0.0003}     & \multicolumn{1}{c}{0.0592$\pm$0.0033}       & \multicolumn{1}{c}{0.0218$\pm$0.0002}        & \multicolumn{1}{c}{0.0271$\pm$0.0048}            & \multicolumn{1}{c}{0.0189$\pm$0.0036}        &  0.2634$\pm$0.0108     \\ \bottomrule
\end{tabular}
\vspace{-10pt}
\end{table*}

\subsection{\name{} as a plug-and-play accelerator for existing simulators}
\textbf{RQ8:} \textit{Can \name{} be applied to existing mobility simulation to improve efficiency while preserving simulation quality?}
\name{} is built to wrap around an existing simulator rather than replace it.
As a case study, we apply \name{} to improve the simulation efficiency of Urban-Mobility-LLM (UML) on the Beijing dataset. Specifically, we first use UML to generate initial mobility trajectory data, which is then fed into our framework for LLM fine-tuning (details in Section~\ref{sec:llm_fine_tuning}). Then we can obtain the required latent embeddings for building the cache. Once the cache is constructed, we can perform efficient mobility simulation.

\begin{figure}[h]\centering
\begin{minipage}[h]{0.47\linewidth}
\vspace{-5pt}
    \includegraphics[width=\linewidth, keepaspectratio=true]{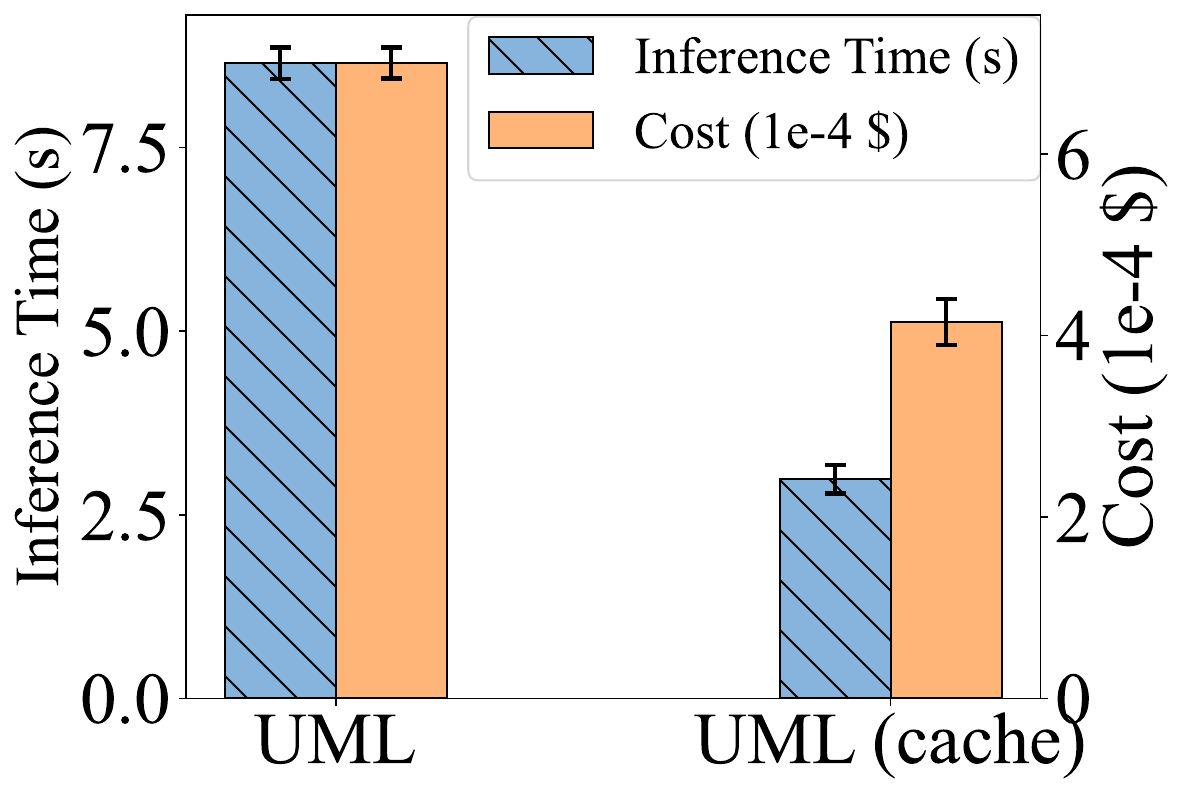}
    \vspace{-10pt}
    \captionsetup{
    font={small},
    skip=2pt  
}
    \caption{Efficiency comparison (UML vs. UML (cache)).}
    \label{fig:case_efficiency}
\end{minipage}
\hspace{5pt}
\begin{minipage}[h]{0.47\linewidth}
\vspace{-5pt}
    \includegraphics[width=\linewidth, keepaspectratio=true]{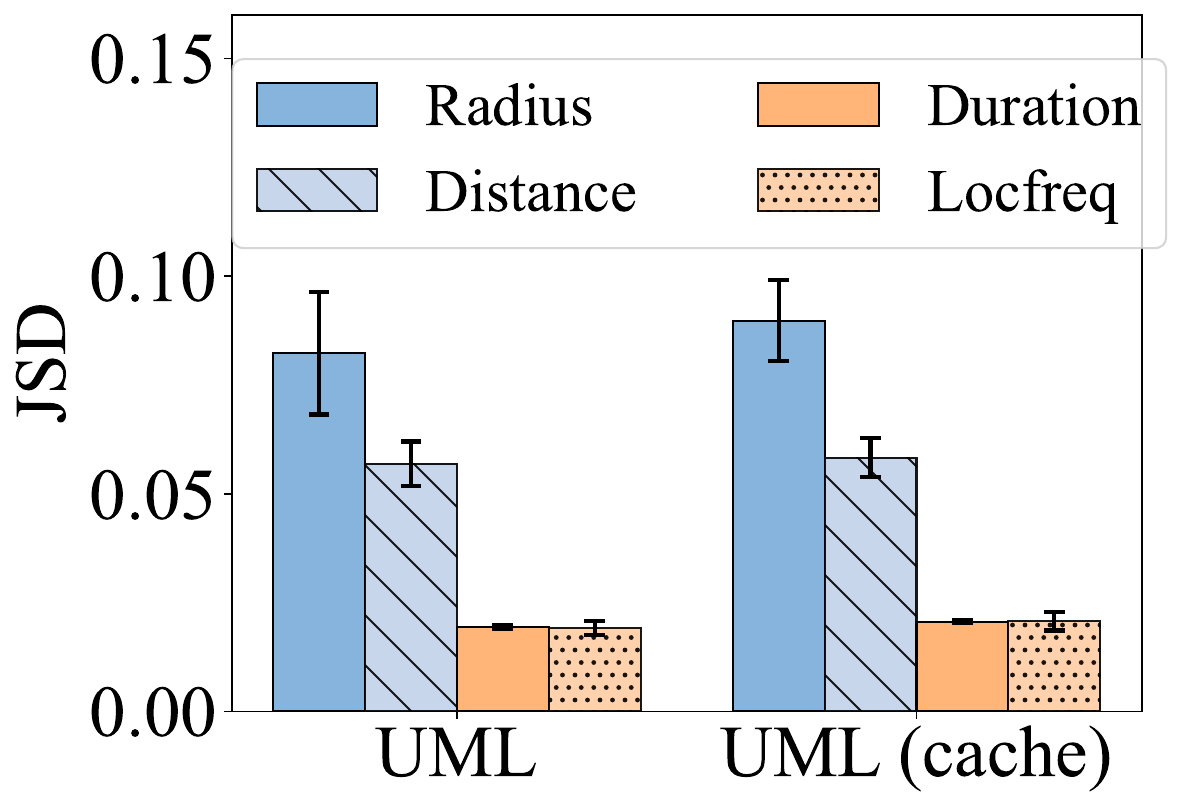}
    \vspace{-10pt}
    \captionsetup{
    font={small},
    skip=2pt  
}
    \caption{Quality comparison (UML vs. UML (cache)).}
    \label{fig:case_performance}
\end{minipage}

\end{figure}

We compare UML and its cache-enhanced version (i.e., UML (cache)), accelerated by our \name{} framework.
In terms of efficiency, as shown in Figure~\ref{fig:case_efficiency}, UML (cache) achieves a simulation speed of 2.985s per trajectory, compared to 8.650s per trajectory  for the original UML, resulting in a 65.49 \% improvement in simulation speed.
In terms of monetary cost, the average expense per trajectory decreases from 
$\$7.00 \times 10^{-4}$ to $\$4.15 \times 10^{-4}$, resulting in a $40.71\%$ improvement in cost efficiency.
In addition, we evaluate quality using four metrics.
As shown in Figure~\ref{fig:case_performance}, the performance of UML (cache) remains comparable to the original UML.

\subsection{Cross-city transferability}
\textbf{RQ9:} \textit{Can a cache constructed from one city be used for mobility simulation in another city?}
We conduct a cross-city transferability experiment to evaluate whether a cache constructed from one city (i.e., Beijing) can effectively accelerate simulations in another city (i.e., New York).
Specifically, we use the cache built from the Beijing dataset in Section 4.3 to accelerate trajectory simulation in New York. We compare \name{} using the NYC cache (i.e., Cache (NYC) for NYC) with \name{} using the Beijing cache (i.e., Cache (BJ) for NYC). For evaluation, we use the NYC POI check-in dataset described in Section~\ref{sec:nyc} as the test set.
It is worth noting that we maintain the same user profile format to enable the retrieval of more similar users from the cache.

\begin{figure}[h]\centering
\begin{minipage}[h]{0.47\linewidth}
    \includegraphics[width=\linewidth, keepaspectratio=true]{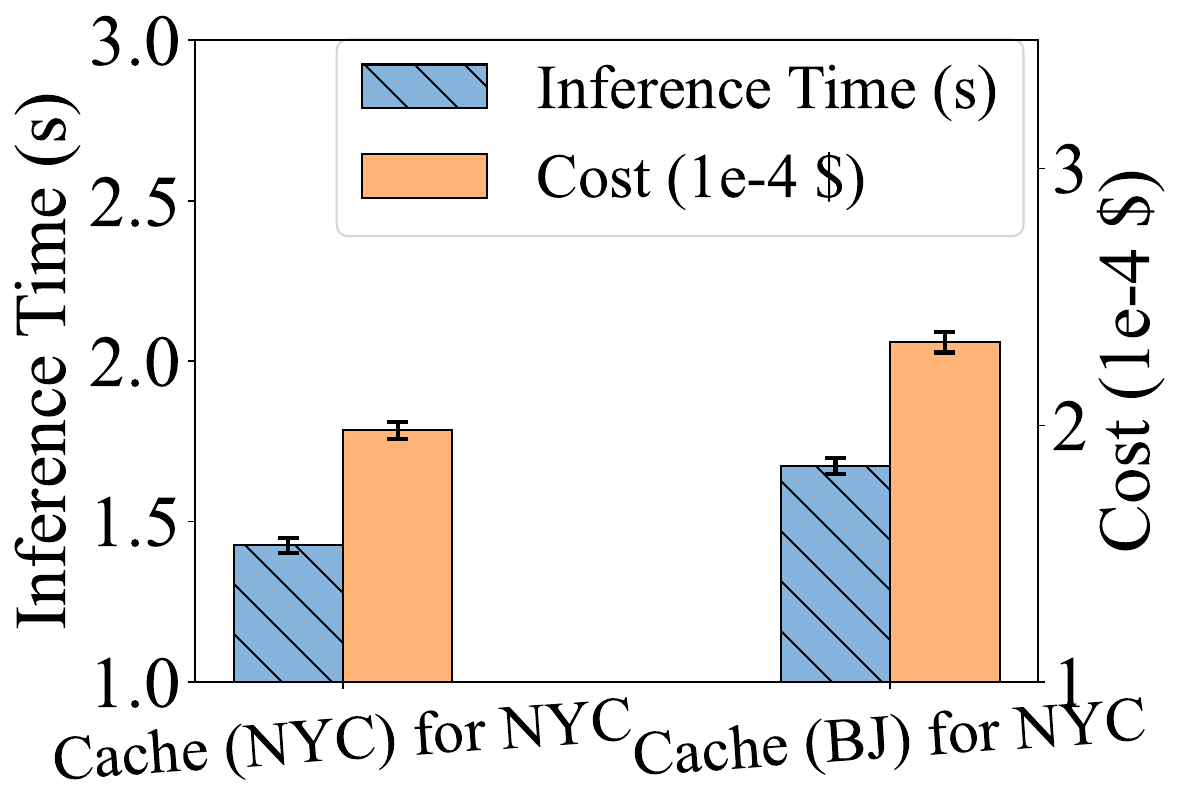}
    \vspace{-10pt}
    \captionsetup{
        font={small},
        skip=2pt  
    }
    \caption{Cross-city: efficiency.}
    \label{fig:trans_efficiency}
\end{minipage}
\hspace{5pt}
\begin{minipage}[h]{0.47\linewidth}
    \includegraphics[width=\linewidth, keepaspectratio=true]{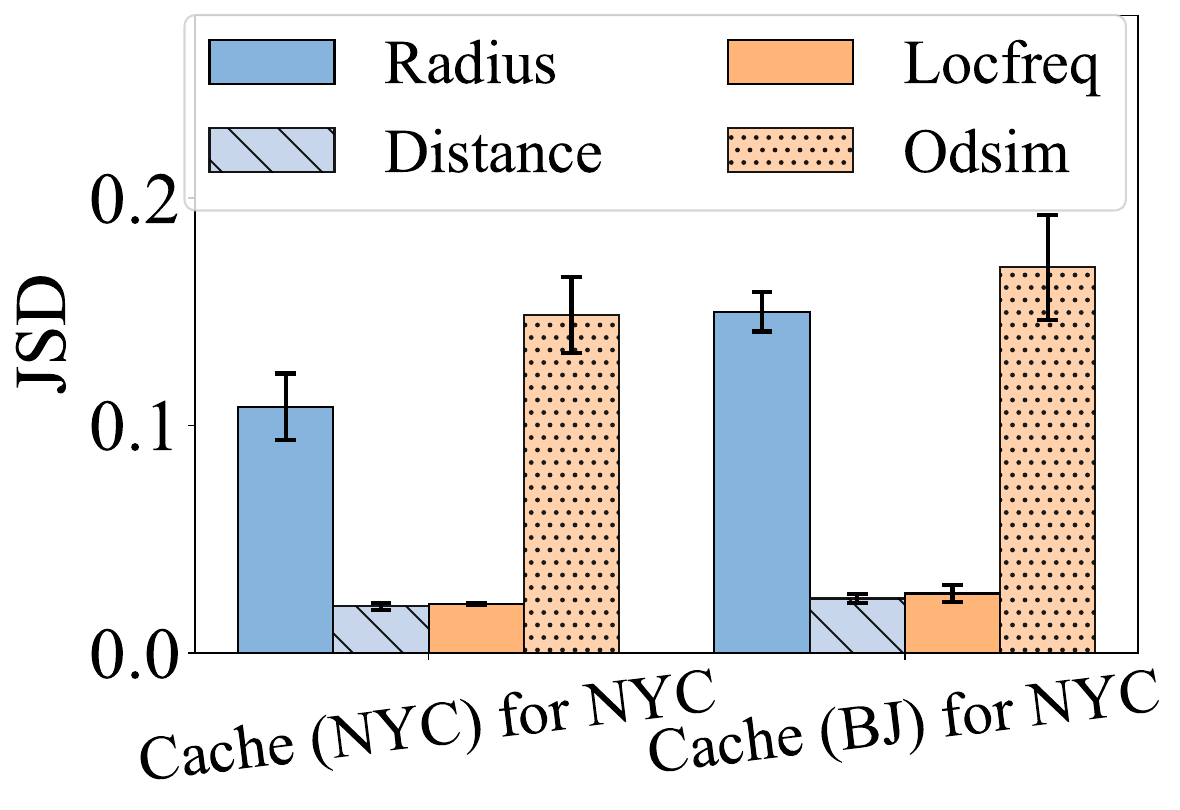}
    \vspace{-10pt}
    \captionsetup{
        font={small},
        skip=2pt  
    }
    \caption{Cross-city: quality.}
    \label{fig:trans_performance}
\end{minipage}
\vspace{-10pt}
\end{figure}

From the Figure~\ref{fig:trans_efficiency}, we observe that, in terms of efficiency, the inference speed on the New York dataset is lower than that on the Beijing dataset. This is because the cache is constructed from Beijing data, resulting in a small portion of New York users failing to find similar matches, which may require additional LLM invocations.
In terms of simulation quality, as shown in Figure~\ref{fig:trans_performance}, the performance is slightly lower than that achieved by a cache specifically built for NYC, but remains acceptable. This is because the cache stores the reasoning process rather than city-specific locations, allowing the reasoning and decision-making to be transferred across cities.

\subsection{Computational cost analysis}
\textbf{RQ10:} \textit{How do the one-time training and cache construction costs of \name{} amortize as the simulation scale increases?}
We analyze end-to-end computational efficiency on the Beijing dataset, measuring total inference time and total cost across simulation scales. To be fair, we fold in all one-time preparation costs: data initialization, model fine-tuning, and cache construction. Because the cache is reused across simulations, these costs are paid once and amortize as the scale grows.
We report the actual cost for the 20K-trajectory setting. For larger scales, we extrapolate from each method's measured per-trajectory cost to estimate city-scale settings that would be too expensive to run in full.

\begin{figure}[h]
\vspace{-5pt}
    \centering
    \includegraphics[width=\linewidth, keepaspectratio=true]{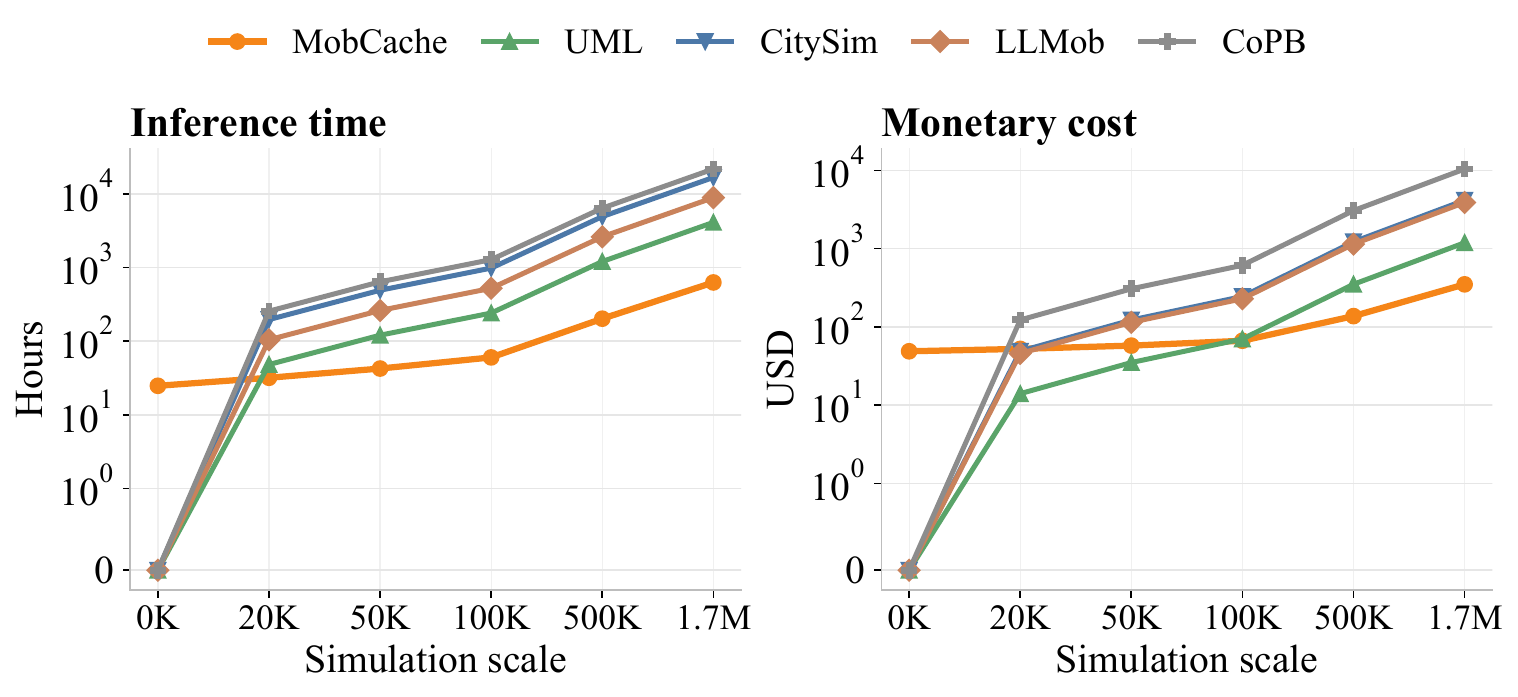}
    \vspace{-15pt}
    \captionsetup{font={small}}
    \caption{Scalability of total inference time and monetary cost across simulation scales, including one-time training costs.}
    \label{fig:time_cost_training}
\vspace{-15pt}
\end{figure}

As shown in Figure~\ref{fig:time_cost_training}, \name{} exhibits substantially better scalability than competing methods in both runtime and monetary cost. Although cache construction introduces additional fixed overhead at smaller scales, this cost is rapidly amortized as the number of simulated trajectories increases. 
At the 100K-trajectory scale, \name{} requires only 60.02 hours and \$66.46, compared with 240.28 hours and \$70.00 for UML, 520.74 hours and \$228.70 for LLMob, and 1274.17 hours and \$612.90 for CoPB. The efficiency gap further widens as the simulation scale increases, demonstrating the effectiveness of cache reuse in reducing repeated LLM inference.

To put the largest simulation scale into context, according to the 2020 United States Census, Manhattan had a resident population of approximately 1.7 million people~\cite{uscensus_quickfacts_nyc}. Under the common assumption of generating one trajectory per resident per day, a realistic one-day city-scale simulation would involve roughly 1.7 million trajectories. 
At this scale, \name{} achieves at least a 6.5× speedup and a 3.4× reduction in monetary cost compared to existing methods. 
These results demonstrate that our framework become increasingly pronounced at realistic urban scales, making large-scale mobility simulation significantly more practical and cost-effective.

\section{Discussion}
\label{sec:Dis}
\textbf{Lessons learned.}
Based on the results from our paper, we summarize the following lessons learned:
\begin{itemize}[leftmargin=*, nosep]
    \item Our reconstructible latent-space cache can augment LLM calls by querying cached results, thereby accelerating large-scale mobility simulation. As shown in Table~\ref{tab:overall}, \name{} achieves comparable performance to state-of-the-art LLM-based methods while significantly reducing inference time and resource consumption.  

    \item Our framework can be generalized to other LLM-based mobility simulation methods. As demonstrated in Figure~\ref{fig:case_efficiency} and Figure~\ref{fig:case_performance}, \name{} significantly enhances the simulation efficiency of UML while maintaining comparable simulation quality.

\end{itemize}

\noindent\textbf{Limitation.} While our framework effectively accelerates other mobility simulation models, it only applies to models that expose interpretable reasoning steps. In particular, the simulation model must provide accessible step-by-step reasoning, either as text or structured latent representations.

\noindent\textbf{Ethics and privacy.} 
This work focuses on accelerating large-scale mobility simulation rather than individual tracking or prediction. Our approach does not require real trajectory data for supervised model training; instead, real-world data is used only for evaluation. Moreover, all trajectory data used for evaluation is fully anonymized and does not contain identifiable user information.
Our framework relies on user profiles from census statistics or publicly available data sources. All information is highly anonymized and represented at a coarse-grained level, preventing the identification of specific individuals and preserving user privacy.
\section{Related work}
\subsection{Human mobility simulation.}
\subsubsection{\textbf{Deep learning based human mobility simulation.}}
Deep learning-based approaches generate mobility trajectories by learning patterns from historical data. Existing studies mainly follow two directions: sequential prediction, which uses RNNs, LSTMs, and Transformers to model spatio-temporal dependencies~\cite{feng2018deepmove, yang2020location, luo2021stan, xu2022metaptp}, and generative simulation, which uses GANs, VAEs, and diffusion models to synthesize trajectories from learned distributions~\cite{feng2020learning, zhu2023difftraj, gupta2018social}. These methods are efficient for large-scale generation once trained.
However, their performance heavily depends on access to large-scale historical mobility datasets. In data-scarce settings, these models often struggle to generalize across new cities, populations, and behavioral contexts. 
Therefore, we do not directly compare with these methods, as they assume access to large-scale real trajectory data for training, whereas such data are not available in our setting. Nevertheless, we provide an experimental comparison in Section~\ref{sec:eval-fine-tune-llm} to illustrate the relative performance of our approach against these methods.

\subsubsection{\textbf{LLM-based human mobility simulation.}}
LLM-based human mobility simulation. Existing works explored using LLMs for human mobility simulation by leveraging their knowledge and human-like reasoning capabilities. The advantage of these approaches is that they do not require large-scale real-world mobility data for training.
These works~\cite{jiawei2024large,du2025cams,piao2025agentsociety,mou2024individual,liu2024human,ju2025trajllm,shao2024chain,bhandari2024urban,li2024geo} guide LLMs to simulate human-like mobility intention reasoning step by step and then produce realistic mobility activity sequences. 
For example, CoPB~\cite{shao2024chain} is an intention and planing based framework that enables LLMs to generate human mobility trajectories through step-by-step reasoning. 
Although these works produce realistic outputs, the reliance on LLMs makes the simulation expensive.
This work~\cite{jiawei2024large} design a LLM-based agent framework for personal mobility generation, combining self-consistency and retrieval strategies to align language models with real-world human activity for accurate and interpretable urban mobility simulation.

To reduce the cost of LLM-based simulation, some works~\cite{chopra2024limits} design a group-based approach, where people are clustered into coarse-grained groups based on profiles and the LLM is invoked once per group to generate shared mobility activities. However, this approach limits diversity, since people in the same group share identical behaviors.
Other works focus on optimizing the efficiency of API interactions~\cite{yan2024opencity,piao2025agentsociety}. For example, OpenCity ~\cite{yan2024opencity} accelerates LLM-based simulation by combining I/O multiplexing and TCP connection pooling to parallelize LLM requests.
AgentSociety~\cite{piao2025agentsociety} accelerates GPT API calls by using Agent Grouping, Ray with asyncio for asynchronous distributed execution, MQTT-based message reuse, and a unified interface for local and remote models, enabling large-scale agent simulations on commodity hardware.
However, the requirement of querying the LLM API for each agent at every simulation step remains, resulting in high cumulative cost.

\subsection{Caches for LLMs.}
When referring to “cache” in the context of LLMs, the most common form is key-value (KV) caching.
KV cache stores the intermediate attention states of previously processed tokens, allowing faster autoregressive decoding without re-computing hidden states. 
Techniques such as prefix caching \cite{kwon2023efficient,liu2024optimizing,liu2023cachegen} reuse KV pairs for initial prompt tokens, while full KV~\cite{gim2024prompt} reuse extends this idea to non-prefix positions via positional embedding shifts.
Our method is orthogonal to traditional KV caching. While KV cache accelerates decoding at the token level within a fixed prompt, our latent cache operates at a higher level by storing latent reasoning steps. This enables flexible reuse across different simulation queries. Importantly, the two approaches are complementary. Our method can potentially benefit further by incorporating KV cache for additional decoding speedup.
In addition, several existing methods leverage query similarity to cache or retrieve useful information, thereby improving efficiency or simulation quality in LLMs~\cite{packer2023memgpt,borgeaud2022improving,lewis2020retrieval}. 
For example, MemGPT~\cite{packer2023memgpt} introduces a memory system that simulates long-term memory, retrieving context dynamically during extended interactions. 
Unlike retrieval methods based on text similarity, our method enables latent-space reasoning reuse, supporting efficient and compositional simulation.
\section{Conclusion}
We presented \name{}, a caching framework that makes large-scale LLM-based mobility simulation practical.
Its central idea is to cache the \emph{reasoning} behind mobility behavior rather than the generated trajectories, and to do so in latent space rather than in language.
Caching reasoning lets a small set of cached chains be recombined into many distinct trajectories, which preserves population diversity; keeping that reasoning in latent space lets us enforce the spatial and temporal constraints that are easily broken when reasoning steps are recombined as text.
\name{} realizes this idea with a reconstructible cache of latent reasoning embeddings, searched as a tree for flexible reuse, and a lightweight decoder, distilled under mobility-law constraints, that turns a reconstructed chain back into a trajectory without repeated calls to the original LLM.
Across two datasets, \name{} matches or exceeds prior LLM-based simulators on trajectory quality while running several times faster and cheaper, and the gap widens at city scale.
Because the cache stores reasoning rather than city-specific locations, it transfers to a new city and scales to larger populations and longer horizons with no retraining, and it can be dropped into an existing simulator to accelerate it without changing the host method.

\newpage
\balance
\bibliographystyle{ACM-Reference-Format}
\bibliography{sample-base}
\newpage
\cleardoublepage
\appendix
\section{Appendix}

\subsection{Baseline adaptation details}
\label{app:baselines}
We provide additional details on how each baseline method is adapted to our data setting to ensure a fair comparison. To further ensure comparability, all baseline methods use GPT-4o-mini as the underlying LLM.

\textbf{CoPB:} We replaced their fine-tuned LLM with an API-based LLM model for fair comparison with other methods, while preserving CoPB’s step-by-step intention reasoning framework. We also reformulated the input profiles and daily records according to the characteristics of the Beijing and NYC datasets.

\textbf{CitySim:} We adapt CitySim to Beijing and NYC datasets by grounding each simulated person in their demographic profile, home location, and observed visit history. The model then generates daily activity trajectories across the user’s observed dates, carrying memory and reflections from previous days to make later simulations more consistent with that person’s routines.

\textbf{UML:} We implement the framework by retaining the 19 activity categories defined in the original paper and using a gravity model to assign concrete locations to each inferred activity within our study area.

\textbf{LLMob:} We adapt LLMob to our data setting by replacing real historical trajectories with a self-growing generated memory: each simulated day is stored and retrieved as behavioral context for subsequent days, avoiding direct use of ground-truth mobility history during generation. Phase 2 is further extended to use this memory for motivation inference and trajectory planning, with explicit plan-level and activity-level rationales while preserving the original pattern-motivation-plan framework.

\subsection{Prompt example}
\begin{tcolorbox}[width=\linewidth, title=Daily mobility activity generation prompt example, label=box:mobility-prompt]
\textbf{Profile}\par
\texttt{Profile:\{profile\}};
\texttt{Date:\{today\_date\}};
\texttt{Nearby Home POIs:\{home\_poi\}};
\texttt{Nearby Home Work:\{work\_poi\}};

\textbf{Task:}
Generate the person's mobility activities for the full day based on profile and home and workplace. Each activity must involve physical movement and staying at a new location.

\textbf{Requirements}\par
\begin{itemize}
    \item Each activity = one meaningful mobility event: movement to a place + stay + purpose.
    \item Prefer activities with spatial diversity: different types of locations and distances (short, medium, long).
    \item Choose realistic activities based on POIs (e.g., gym near home, restaurant near work, hospital visit, etc).
    \item Encourage inclusion of diverse behavior types: health (e.g., walk, gym), errands, social, entertainment, unusual events.
    \item The person’s day must have between 2 and 9 total mobility activities.
    \item Output two sections: \texttt{Reasoning} and \texttt{Final Activities}.
\end{itemize}

\textbf{Output Format}\par
\texttt{Reasoning:}
\begin{itemize}
    \item At 12:30 a.m., After late night out, he may go home for sleep. Distance: 8km.
    \item ...
\end{itemize}

\texttt{Final Activities:} 1. At 12:30 a.m., Return home, 8km... 
\end{tcolorbox}

\clearpage
\twocolumn[{%
\begin{minipage}{\dimexpr(\textwidth-\columnsep)/2\relax}
\subsection{Notation}
For clarity, Table~\ref{tab:notations} summarizes the main notations used throughout the paper,
including the latent-space reasoning process, reconstructible cache,
latent-space evaluator, and lightweight decoder.
\end{minipage}

\vspace{8pt}

\begin{center}
\captionof{table}{Summary of notations.}
\label{tab:notations}

\small
\setlength{\tabcolsep}{4pt}
\renewcommand{\arraystretch}{1.0}

\begin{tabular}{
@{}
>{\raggedright\arraybackslash}p{0.18\textwidth}
>{\raggedright\arraybackslash}p{0.74\textwidth}
@{}
}
\toprule
\textbf{Notation} & \textbf{Description} \\
\midrule
$q$ & Input prompt containing user profile, date, POIs, and task-specific information. \\
$y$ & Final mobility activity output. \\
$Y=(y_1,\ldots,y_L)$ & Generated mobility activity sequence. \\
$L$ & Length of the activity sequence. \\
$r_t$ & Latent-space reasoning embedding at reasoning step $t$. \\
$\hat r_t$ & Next latent reasoning embedding generated by the latent reasoning model during evaluator training. \\
$R_{1:t}$ &  Latent reasoning chain $[r_1,r_2,\ldots,r_t]$. \\
$f_\theta(\cdot)$ & Fine-tuned latent reasoning model. \\
$g_\xi(\cdot)$ & Latent-space evaluator. \\
$\sigma_t$ & Evaluator score assigned to candidate reasoning step $r_t$. \\
$b_t$ & Supervision label used for evaluator training. \\
$\mathrm{sim}(\cdot,\cdot)$ & Similarity function between latent reasoning embeddings. \\
$e_\mu(\cdot)$ & MLP projector that maps teacher latent embeddings into the lightweight decoder space. \\
$\hat R$ & Projected latent reasoning chain used as decoder input. \\
$P_\rho(\cdot)$ & Output distribution of the lightweight decoder. \\
$h_{\text{light}}$ & Hidden representation of the lightweight decoder. \\
$z_\tau(\cdot)$ & Mobility-law prediction network. \\
$d(\cdot)$ & Function extracting mobility-law statistics from activity sequences. \\
$p_{\text{teacher}}(d(Y))$ & Mobility-law distribution derived from activities generated by the teacher decoder. \\
$L_{\text{distill}}$ & Distillation loss for lightweight decoder training. \\
$L_{\text{law}}$ & Mobility-law constraint loss. \\
$L_{\text{total}}$ & Overall training objective. \\
$\lambda$ & Weight balancing the distillation loss and mobility-law constraint loss. \\
\bottomrule
\end{tabular}
\end{center}

\vspace{10pt}
}]

\end{document}